%% file: paper.tex
\documentclass{article}

\PassOptionsToPackage{numbers}{natbib}

\newif\ifpreprint

\preprinttrue  %
\usepackage[final]{lib/neurips_2022}  %
\input{lib/preamble}

\input{lib/macros}
\input{lib/math_macros}

\title{Beyond spectral gap:\\ The role of the topology in decentralized learning}

\author{%
  Thijs Vogels\thanks{Equal contribution. Corresponding authors \texttt{thijs.vogels@epfl.ch} and \texttt{hadrien.hendrikx@epfl.ch}.} \\
  EPFL
  \And
  Hadrien Hendrikx$^*$  \\
  EPFL
  \And
  Martin Jaggi \\
  EPFL
}

\begin{document}

\maketitle

\input{010_abstract}
\input{020_introduction}
\input{030_related_work}
\input{040_random_quadratics}
\input{060_theory}
\input{065_experiments}
\input{070_conclusion}
\input{080_acknowledgements}

\newpage

\bibliography{bibliography_autogen}

\ifpreprint
\else
	\input{090_checklist}
\fi

\newpage
\input{095_appendix}

\end{document}

%% file: lib/preamble.tex
\usepackage[utf8]{inputenc} 
\usepackage[T1]{fontenc}    
\usepackage{microtype}      
\usepackage{xspace}         
\usepackage{ifthen}

\usepackage{xargs}          

\usepackage{amssymb,amsmath,amsthm,amsfonts}
\usepackage{bm}             
\usepackage{mathtools}
\usepackage{dsfont}

\usepackage{algorithm}
\usepackage{algpseudocode}  
\usepackage{booktabs}
\usepackage{tabularx}       

\usepackage[dvipsnames]{xcolor} 
\usepackage{graphicx}
\usepackage{tikz}           

\usetikzlibrary{patterns}
\usetikzlibrary{positioning}

\usepackage[toc, page, header]{appendix}
\usepackage{url}

\usepackage{hyperref}       
\hypersetup{hidelinks}

%% file: lib/macros.tex
\newcommand{\eg}{e.g.\xspace}
\newcommand{\ie}{i.e.\xspace}
\newcommand{\iid}{i.i.d.\ }

\newcommand{\nocolor}[1]{}

\newcommand{\dims}{{{\nocolor{tab10_red}d}}}
\newcommand{\nwrk}{{{\nocolor{tab10_purple}n}}}
\newcommand{\lr}{{{\nocolor{tab10_red}\eta}}}
\newcommand{\gossip}{{{\nocolor{tab10_purple}\mW}}}
\newcommand{\weigval}{{{\nocolor{tab10_purple}\lambda}}}
\newcommand{\weigvec}{{{\nocolor{tab10_purple}\vv}}}
\newcommand{\gossipweight}[1]{{{\nocolor{tab10_purple}w_{#1}}}}
\newcommand{\param}{{{\nocolor{tab10_orange}\xx}}}
\newcommand{\grad}{{{\nocolor{tab10_orange}\gg}}}
\newcommand{\trans}{{{\nocolor{tab10_cyan}\mT}}}
\newcommand{\noise}{{{\nocolor{tab10_green}\zeta}}}
\newcommand{\smoothf}{{{\nocolor{tab10_blue}L}}}
\newcommand{\randdata}{{{\nocolor{tab10_blue}\dd}}}
\newcommand{\variance}{{{\nocolor{tab10_green}\sigma}}}
\newcommand{\mmat}{{{\nocolor{tab10_olive}\mM}}}
\newcommand{\mweight}[1]{{{\nocolor{tab10_olive}\mmat_{#1}}}}
\newcommand{\rate}{{{\nocolor{tab10_brown}r}}}
\newcommand{\LM}{{{\nocolor{tab10_pink}\mL_\mmat}}}
\newcommand{\LW}{{{\nocolor{tab10_pink}\mL_{\mW}}}}
\newcommand{\wcons}{{{\nocolor{tab10_gray}\omega}}}
\newcommand{\decay}{{{\nocolor{tab10_cyan}\gamma}}}

\newcommand{\mvar}{{{\nocolor{tab10_purple}\tilde{\variance}^2}}}
\newcommand{\neff}{{\nocolor{tab10_olive}{n_\gossip(\decay)}}}

\newcommand{\lmax}{{{\nocolor{tab10_orange}\weigval_2}}}

\newcommand{\esp}[1]{\expect\left[#1\right]}

\newcommand{\rw}{{{\nocolor{tab10_orange}\zz}}}
\newcommand{\rwo}{{{\nocolor{tab10_orange}\yy}}}
\newcommand{\rwn}{{{\nocolor{tab10_orange}\bm{\xi}}}}
\newcommand{\rwcov}{{{\nocolor{tab10_orange}\mC}}}
\newcommand{\rwcovv}{{{\nocolor{tab10_orange}\cc}}}
\newcommand{\rwcoveig}{{{\nocolor{tab10_orange}c}}}

\newcommand{\emat}{{{\nocolor{tab10_blue}\mE}}}
\newcommand{\evec}{{{\nocolor{tab10_blue}\ee}}}

\newcommand{\rws}{{{\nocolor{tab10_pink}\yy}}}
\newcommand{\rwcovs}{{{\nocolor{tab10_pink}\mB}}}
\newcommand{\rwcovvs}{{{\nocolor{tab10_pink}\bb}}}

\newcommand{\dsgd}{D-SGD\xspace}

\definecolor{tab10_blue}{rgb}{0.121, 0.466, 0.705}
\definecolor{tab10_orange}{rgb}{1.0,   0.498, 0.054}
\definecolor{tab10_green}{rgb}{0.172, 0.627, 0.172}
\definecolor{tab10_red}{rgb}{0.839, 0.152, 0.156}
\definecolor{tab10_purple}{rgb}{0.580, 0.403, 0.741}
\definecolor{tab10_brown}{rgb}{0.549, 0.337, 0.294}
\definecolor{tab10_pink}{rgb}{0.890, 0.466, 0.760}
\definecolor{tab10_gray}{rgb}{0.498, 0.498, 0.498}
\definecolor{tab10_olive}{rgb}{0.737, 0.741, 0.133}
\definecolor{tab10_cyan}{rgb}{0.090, 0.745, 0.811}

\input{lib/comments}

\newcommand*{\at}[1]{\ifthenelse{\equal{#1}{\star}}{^\star}{^{(#1)}}}
\newcommand*{\idx}[1]{_{#1}}
\newcommand*{\atidx}[2]{\at{#1}\idx{#2}}

%% file: lib/comments.tex
\def\commentType{0}

\ifnum\commentType=0
  \usepackage[disable]{todonotes} 
  \newcommand{\customComment}[3]{\todo[color=#2!40,size=\small]{\textbf{#1:} #3}}
  
\fi
\ifnum\commentType=1
  \newcommand{\customComment}[3]{\textcolor{#2}{\textsl{#1:~#3}}}
  
\fi
\ifnum\commentType=2
  \usepackage{pdfcomment}
  \newcommand{\customComment}[3]{\pdfcomment[icon=Comment,opacity=0.5,color=#2,author=#1]{#3}}
  
\fi
\ifnum\commentType=3
  \usepackage{todonotes} 
  \usepackage{silence}
  \newcommand{\customComment}[3]{\todo[color=#2!40,size=\small]{\textbf{#1:} #3}\xspace}
  
\fi

\newcommand{\Hadrien}[1]{\customComment{Hadrien}{tab10_blue}{#1}}
\newcommand{\Thijs}[1]{\customComment{Thijs}{tab10_orange}{#1}}

%% file: lib/math_macros.tex
\providecommand{\lin}[1]{\left\langle #1 \right\rangle}

\DeclarePairedDelimiterX{\abs}[1]{\lvert}{\rvert}{#1}
\DeclarePairedDelimiterX{\norm}[1]{\lVert}{\rVert}{#1}
\DeclarePairedDelimiterX{\cbr}[1]{\{}{\}}{#1} 
\DeclarePairedDelimiterX{\rbr}[1]{(}{)}{#1} 
\DeclarePairedDelimiterX{\sbr}[1]{[}{]}{#1} 

\DeclareMathOperator{\Var}{Var}

\providecommand{\R}{\mathbb{R}} 

\DeclareMathOperator{\expect}{\mathbb{E}}

\DeclareMathOperator{\sgn}{sign}
\makeatletter
\def\sign{\@ifnextchar*{\@sgnargscaled}{\@ifnextchar[{\sgnargscaleas}{\@ifnextchar{\bgroup}{\@sgnarg}{\sgn} }}}
\def\@sgnarg#1{\sgn\rbr{#1}}
\def\@sgnargscaled#1{\sgn\rbr*{#1}}
\def\@sgnargscaleas[#1]#2{\sgn\rbr[#1]{#2}}
\makeatother
\DeclareMathOperator*{\argmin}{arg\,min}

\DeclareMathOperator*{\Tr}{Tr}

\providecommand{\0}{\mathbf{0}}

\providecommand{\bb}{\mathbf{b}}
\providecommand{\cc}{\mathbf{c}}
\providecommand{\dd}{\mathbf{d}}
\providecommand{\ee}{\mathbf{e}}

\let\ggg\gg
\renewcommand{\gg}{\mathbf{g}}

\providecommand{\vv}{\mathbf{v}}

\providecommand{\xx}{\mathbf{x}}
\providecommand{\yy}{\mathbf{y}}
\providecommand{\zz}{\mathbf{z}}

\providecommand{\mB}{\mathbf{B}}
\providecommand{\mC}{\mathbf{C}}

\providecommand{\mE}{\mathbf{E}}

\providecommand{\mI}{\mathbf{I}}

\providecommand{\mL}{\mathbf{L}}
\providecommand{\mM}{\mathbf{M}}

\providecommand{\mP}{\mathbf{P}}
\providecommand{\mQ}{\mathbf{Q}}

\providecommand{\mT}{\mathbf{T}}

\providecommand{\mW}{\mathbf{W}}

\providecommand{\cL}{\mathcal{L}}

\providecommand{\cN}{\mathcal{N}}
\providecommand{\cO}{\mathcal{O}}

\newtheoremstyle{tstyle}
  {\topsep} 
  {0} 
  {} 
  {} 
  {\bfseries} 
  {.} 
  {.5em} 
  {} 

\theoremstyle{tstyle}
\newtheorem{theorem}{Theorem}

  \newtheorem{corollary}[theorem]{Corollary}

\newtheorem{lemma}{Lemma}

\newtheorem{definition}{Definition}

\newtheorem{assumption}[definition]{Assumption}

\newtheorem*{nonumber_assumption}{Assumption B}

%% file: 010_abstract.tex
\begin{abstract}
	In data-parallel optimization of machine learning models, workers collaborate to improve their estimates of the model: more accurate gradients allow them to use larger learning rates and optimize faster.
	We consider the setting in which all workers sample from the same dataset,
 	and communicate over a sparse graph (decentralized).
	In this setting, current theory fails to capture important aspects of real-world behavior.
	First, the `spectral gap' of the communication graph is not predictive of its empirical performance in (deep) learning.
	Second, current theory does not explain that collaboration enables \emph{larger} learning rates than training alone.
	In fact, it prescribes \emph{smaller} learning rates, which further decrease as graphs become larger, failing to explain convergence in infinite graphs.
	This paper aims to paint an accurate picture of sparsely-connected distributed optimization when workers share the same data distribution.
	We quantify how the graph topology influences convergence in a quadratic toy problem and provide theoretical results for general smooth and (strongly) convex objectives.
	Our theory matches empirical observations in deep learning, and accurately describes the relative merits of different graph topologies.
	Code: \texttt{\href{https://github.com/epfml/topology-in-decentralized-learning}{github.com/epfml/topology-in-decentralized-learning}}
\end{abstract}

%% file: 020_introduction.tex
\section{Introduction}
\label{sec:intro}

Distributed data-parallel optimization algorithms help us tackle the increasing complexity of machine learning models and of the data on which they are trained.
We can classify those training algorithms as either \emph{centralized} or \emph{decentralized}, and we often consider those settings to have different benefits over training `alone'.
In the \emph{centralized} setting, workers compute gradients on independent mini-batches of data, and they average those gradients between all workers.
The resulting lower variance in the updates enables larger learning rates and faster training.
In the \emph{decentralized} setting, workers average their models with only a sparse set of `neighbors' in a graph instead of all-to-all, and they may have private datasets sampled from different distributions.
As the benefit of decentralized learning, we usually focus only on the (indirect) access to other worker's datasets, and not of faster training.

While decentralized learning is typically studied with heterogeneous datasets across workers, sparse (decentralized) averaging between is also useful when worker's data is identically distributed (i.i.d.)~\cite{lu2021optimal}.
As an example, sparse averaging is used in data centers to mitigate communication bottlenecks~\cite{assran2019sgp}.
In fact the \dsgd algorithm~\cite{lian2017can}, on which we focus in this work, performs well mainly in this setting, while algorithmic modifications~\cite{Lorenzo2016GT-first-paper,tang2018d2,vogels2021relay} are required to yield good performance on heterogeneous objectives.
When the data is i.i.d., the goal of sparse averaging is to optimize faster, just like in centralized (all-to-all) graphs.

\begin{figure}
    \includegraphics{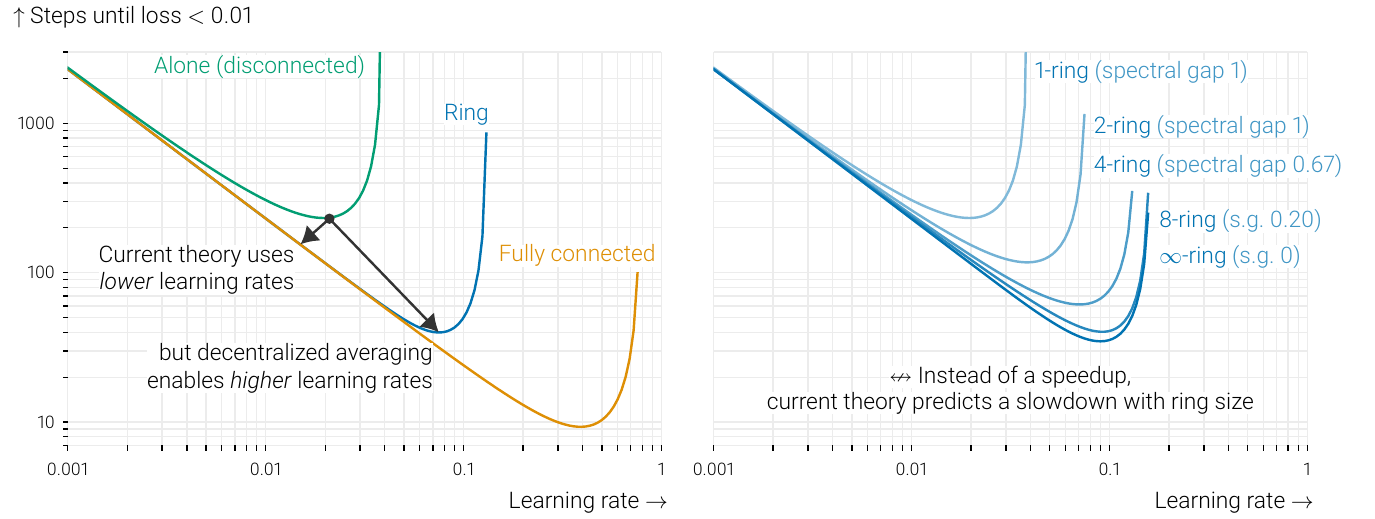}
    \centering
    \caption{
        \label{fig:intro_illustration}
        `Time to target' for \dsgd~\cite{lian2017can} with constant learning rates on an \iid isotropic quadratic dataset (\autoref{sec:toy-model}).
        The noise disappears at the optimum.
        Compared to optimizing alone, 32 workers in a ring (\emph{left}) are faster for any learning rate, but the largest improvement comes from being able to use a large learning rate.
        This benefit is not captured by current theory, which prescribes a smaller learning rate than training alone.
        On the \emph{right}, we see that rings of increasing size enable larger learning rates and faster optimization.
        Because a ring's spectral gap goes to zero with the size, this cannot be explained by current theory.
    }
\end{figure}

Yet, current decentralized learning theory poorly explains the \iid case.
Analyses typically show that, for \emph{small enough} learning rates, training with sparse averaging behaves the same as with all-to-all averaging~\citep{lian2017can,koloskova2020unified}.
Compared to training alone with the \emph{same small learning rate}, all-to-all averaging reduces the gradient variance by the number of workers.
In practice, however, such small learning rates would never be used.
In fact, a reduction in variance should allow us to use a \emph{larger} learning rate than training alone, rather than imposing a \emph{smaller} one.
Contrary to current theory, we show that averaging reduces the variance from the start, instead of just asymptotically.
Lower variance increases the maximum learning rate, which directly speeds up convergence.
We characterize how much averaging with various communication graphs reduces the variance, and show that centralized performance is not always achieved when using optimal large learning rates.
The behavior we explain is illustrated in \autoref{fig:intro_illustration}.

In current convergence rates, the graph topology appears through the \emph{spectral gap} of its averaging (gossip) matrix.
The spectral gap poses a conservative lower bound on how much one averaging step brings all worker's models closer together.
The larger, the better.
If the spectral gap is small, a significantly smaller learning rate is required to make the algorithm behave close to SGD with all-to-all averaging with the same learning rate.
Unfortunately, we experimentally observe that, both in deep learning and in convex optimization, the spectral gap of the communication graph is \emph{not predictive} of its performance under realistically tuned learning rates.

The problem with the spectral gap quantity is clearly illustrated in a simple example.
Let the communication graph be a ring of varying size.
As the size of the ring increases to infinity, its spectral gap goes to zero, since it becomes harder and harder to achieve consensus between all the workers.
This leads to the optimization progress predicted by current theory to go to zero as well.
Yet, this behavior does not match the empirical behavior of the rings with \iid data.
As the size of the ring increases, the convergence rate actually \emph{improves}~(\autoref{fig:intro_illustration}), until it saturates at a point that depends on the problem.

In this work, we aim to  aaccurately describe the behavior of \iid distributed learning algorithms with sparse averaging, both in theory and in practice.
We quantify the role of the graph in a quadratic toy problem designed to mimic the initial phase of deep learning (\autoref{sec:toy-model}), showing that averaging enables a larger learning rate.
From these insights, we derive a problem-independent notion of `effective number of neighbors' in a graph that is consistent with time-varying topologies and infinite graphs, and is predictive of a graph's empirical performance in both convex and deep learning.
We provide convergence proofs for convex and (strongly) convex objectives that only mildly depend on the spectral gap of the graph (\autoref{sec:theory}), and consider the whole spectrum instead.
At its core, our analysis does not enforce global consensus, but only between workers that are close to each other in the graph.
Our theory shows that sparse averaging provably enables larger learning rates and thus speeds up optimization.
These insights prove to be relevant in deep learning, where
we accurately describe the performance of a variety of topologies, while their spectral gap does not (\autoref{sec:experiments}).

%% file: 030_related_work.tex
\section{Related work}
\label{sec:related_work}

\vspace{-.2mm}
\paragraph{Decentralized SGD}
This paper studies decentralized SGD.
\citet{koloskova2020unified} obtain the tightest bounds for this algorithm in the general setting where workers optimize heterogeneous objectives.
Contrary to their work, we focus primarily on the case where all workers sample \iid data from the same distribution.
This important case is not described in a meaningful way by their analysis:
while they show that gossip averaging reduces the asymptotic variance suffered by the algorithm, the fast initial linear decrease term in their convergence rate depends on the spectral gap of the gossip matrix.
This key term does not improve through collaboration and gives rise to a \emph{smaller learning rate} than training alone.
Besides, as discussed above, this implies that optimization is not possible in the limit of large graphs, even in the absence of heterogeneity: for instance, the spectral gap of an infinite ring is zero, which would lead to a learning rate of zero as well.

These rates suggest that decentralized averaging speeds up the last part of training (dominated by variance), at the cost of slowing down the initial (linear convergence) phase.
Beyond the work of~\citet{koloskova2020unified}, many papers focus on \emph{linear speedup} (in the variance phase) over optimizing alone, and prove similar results in a variety of settings~\citep{lian2017can,tang2018d,lian2018asynchronous}.
All these results rely on the following insight: while linear speedup is only achieved for small learning rates, SGD eventually requires such small learning rates anyway (because of, \eg, variance, or non-smoothness).
This observation leads these works to argue that ``topology does not matter''.
This is the case indeed, but only for very small learning rates, as shown in \autoref{fig:intro_illustration}.
In practice, averaging speeds up both the initial \emph{and} last part of training.
This is what we show in this work, both in theory and in practice.

Another line of work studies D-(S)GD under statistical assumptions on the local data.
In particular, \citet{richards2020graph} show favorable properties for \dsgd with graph-dependent implicit regularization and attain optimal statistical rates.
Their suggested learning rate does depend on the spectral gap of the communication network, and it goes to zero when the spectral gap shrinks.
\citet{richards2019optimal} also show that larger (constant) learning rates can be used in decentralized GD, but their analysis focuses on decentralized kernel regression. It does not cover stochastic gradients, and relies on statistical concentration of local objectives rather than analysis on local neighborhoods.

\vspace{-.2mm}
\paragraph{Gossiping in infinite graphs}
An important feature of our results is that they only mildly depend on the spectral gap, and so they apply independently of the size of the graph.
\citet{berthier20acceleratedgossip} study acceleration of gossip averaging in infinite graphs, and obtain the same conclusions as we do:
although spectral gap is useful for asymptotics, it fails to accurately describe the transient regime of averaging.
This is especially limiting for optimization (compared to of just averaging), as new local updates need to be averaged at every step.
The transient regime of averaging deeply matters.
Indeed, it impacts the quality of the gradient updates, and so it rules the asymptotic regime of optimization.

\vspace{-.2mm}
\paragraph{The impact of the topology}
Some works on linear speedup~\citep{lian2017can} argue that the topology of the graph does not matter.
This is only true for asymptotic rates in specific settings, as illustrated in \autoref{fig:intro_illustration}.
\citet{neglia20doestopomatter} investigate the impact of the topology on decentralized optimization, and contradict this claim.
Compared to us, they make different noise assumptions, which in particular depend on the spectral distribution of the noise over the eigenvalues of the Laplacian (thus mixing computation and communication aspects).
Although they show that the topology has an impact in the early phases of training (just like we do), they still get an unavoidable dependence on the spectral gap of the graph.
Our results are different in nature, and show the benefits of averaging and the impact of the topology through the choice of large learning rates.

Another line of work studies the interaction of topology with particular patterns of data heterogeneity~\citep{dandi2022data,lebars2022yes}, and how to optimize graphs with this heterogeneity in mind.
These works ``only'' show a benefit from one-step gossip averaging and this is thus what they optimize the graph for.
In contrast, we show that it is possible to benefit from distant workers beyond direct neighbors, too.
This is an orthogonal direction, though the insights from our work could be used to strengthen their results.

\vspace{-.2mm}
\paragraph{Time-varying topologies}
Time-varying topologies are popular for decentralized deep learning in data centers due to their strong mixing~\cite{assran2019sgp,wang2019matcha}.
The benefit of varying the communication topology over time is not easily explained through standard theory, but requires dedicated analysis~\cite{ying2021exponential}.
While our proofs only cover static topologies, the quantities that appear in our analysis can be computed for time-varying schemes, too.
With these quantities, we can empirically study static and time-varying schemes in the same framework.

%% file: 040_random_quadratics.tex
\section{A toy problem: \dsgd on isotropic random quadratics}
\label{sec:toy-model}

Before analyzing decentralized stochastic optimization through theory for general convex objectives and deep learning experiments, we first investigate a simple toy example that illustrates the behavior we want to explain in the analysis.
In this setting, we can exactly characterize the convergence of decentralized SGD.
We also introduce concepts that will be used throughout the paper.

We consider $\nwrk$ workers that jointly optimize an isotropic quadratic $\expect_{\randdata\sim \cN^\dims(0,1)} \frac{1}{2}(\randdata^\top \param)^2 = \frac{1}{2}\norm{\param}^2$ with a unique global minimum $\param^\star = \0$.
The workers access the quadratic through stochastic gradients of the form $\grad(\param) = \randdata \randdata^\top \param$, with $\randdata \sim \cN^\dims(0, 1)$.
This corresponds to a linear model with infinite data, and where the model can fit the data perfectly, so that stochastic noise goes to zero close to the optimum.
We empirically find that this simple model is a meaningful proxy for the initial phase of (over-parameterized) deep learning (\autoref{sec:experiments}).
A benefit of this model is that we can compute exact rates for it. These rates illustrate the behavior that we capture more generally in the theory of \autoref{sec:theory}.
\autoref{apx:random-quadratics} contains a detailed version of this section that includes full derivations.

The stochasticity in this toy problem can be quantified by the \emph{noise level}
\begin{align}
    \noise = \sup_{\param \in \R^d} \frac{\expect_\randdata \norm{\randdata \randdata^\top \param}^2}{\norm{\param}^2},
\end{align}
which is equal to $\noise=\dims+2$, due to the random normal distribution of $\randdata$.

The workers run the \dsgd algorithm~\cite{lian2017can}.
Each worker $i$ has its own copy $\param\idx{i} \in \R^\dims$ of the model, and they alternate between local model updates $\param\idx{i} \gets \param\idx{i} - \lr\mspace{1mu}\grad(\param\idx{i})$ and averaging their models with others: $\param\idx{i} \gets \sum_{j=1}^\nwrk \gossipweight{ij} \param\idx{j}$.
The averaging weights $\gossipweight{ij}$ are summarized in the \emph{gossip matrix} $\gossip \in \R^{\nwrk\times \nwrk}$.
A non-zero weight $\gossipweight{ij}$ indicates that $i$ and $j$ are directly connected.
In the following, we assume that $\gossip$ is symmetric and doubly stochastic: $\sum_{j=1}^n \gossipweight{ij} = 1\;\forall i$.

On our objective, \dsgd either converges or diverges linearly.
Whenever it converges, \ie when the learning rate is small enough, there is a convergence rate $\rate$ such that \vspace{-1mm}
\begin{align*}
    \expect \norm{\param\atidx{t}{i}}^2 \leq (1 - \rate) \norm{\param\atidx{t-1}{i}}^2,
\end{align*}
with equality as $t \to \infty$ (proofs in~\autoref{apx:random-quadratics}).
When the workers train alone ($\gossip = \mI$), the convergence rate for a given learning rate $\lr$ reads: \vspace{-1mm}
\begin{align}
    \label{eq:solo}
    \rate_\text{alone} = 1 - {\color{tab10_red}(1 - \lr)^2} - {\color{tab10_green}(\noise - 1) \lr^2}.
\end{align}\vspace{-1mm}%
The optimal learning rate $\lr^\star = \frac{1}{\noise}$ balances the optimization term $\color{tab10_red}(1-\lr)^2$ and the stochastic term $\color{tab10_green}(\noise - 1) \lr^2$.
In the centralized (fully connected) setting ($\gossipweight{ij} = \frac{1}{\nwrk}\; \forall i,j$), the rate is simple as well:\vspace{-2mm}
\begin{align}
    \label{eq:centralized}
    \rate_\text{centralized} = 1 - {\color{tab10_red}(1 - \lr)^2} - \frac{\color{tab10_green}(\noise - 1) \lr^2}{\color{tab10_blue} \nwrk}.
\end{align}
Averaging between $\color{tab10_blue}n$ workers reduces the impact of the gradient noise, and the optimal learning rate grows to $\lr^\star = \frac{\color{tab10_blue}\nwrk}{{\color{tab10_blue}\nwrk}+\noise - 1}$.
\dsgd with a general gossip matrix $\gossip$ interpolates those results.

To quantify the reduction of the $\color{tab10_green}(\noise - 1) \lr^2$ term in general, we introduce the \emph{problem-independent} notion of \emph{effective number of neighbors} $\nwrk_\gossip(\decay)$ of the gossip matrix $\gossip$ and \emph{decay parameter} $\decay$.

\begin{definition}
    The effective number of neighbors $\nwrk_{\color{tab10_blue}\gossip}({\decay}) = \lim_{t\to\infty}  \frac{\sum_{i=1}^\nwrk\Var[\rwo\at{t}\idx{i}]}{\sum_{i=1}^\nwrk\Var[\rw\at{t}\idx{i}]}$ measures the ratio of the asymptotic variance of the processes
    \begin{align}
        \label{eqn:rw-noavg}
        \rwo\at{t+1} =\sqrt{{\decay}} \cdot \rwo\at{t} + \rwn\at{t},
        \quad \text{where } \rwo\at{t} \in \R^\nwrk \text{ and } \rwn\at{t} \sim \cN^\nwrk(0,1)
    \end{align}
    and
    \begin{align}
        \label{eqn:rw}
        \rw\at{t+1} = {\color{tab10_blue}\gossip}( \sqrt{{\decay}} \cdot \rw\at{t} + \rwn\at{t}),
        \quad \text{where } \rw\at{t} \in \R^\nwrk \text{ and } \rwn\at{t} \sim \cN^\nwrk(0,1).
    \end{align}
\end{definition}
We call $\rwo$ and $\rw$ \emph{random walks} because workers repeatedly add noise to their state, somewhat like SGD's parameter updates.
This should not be confused with a `random walk' over nodes in the graph.

Since averaging with ${\color{tab10_blue}\gossip}$ decreases the variance of the random walk by at most $\nwrk$, the effective number of neighbors is a number between $1$ and $\nwrk$.
The decay $\decay$ modulates the sensitivity to communication delays.
If $\decay=0$, workers only benefit from averaging with their direct neighbors.
As $\decay$ increases, multi-hop connections play an increasingly important role.
As $\decay$ approaches 1, delayed and undelayed noise contributions become equally weighted, and the reduction tends to $\nwrk$ for any connected topology.

\begin{figure}
    \centering
    \includegraphics{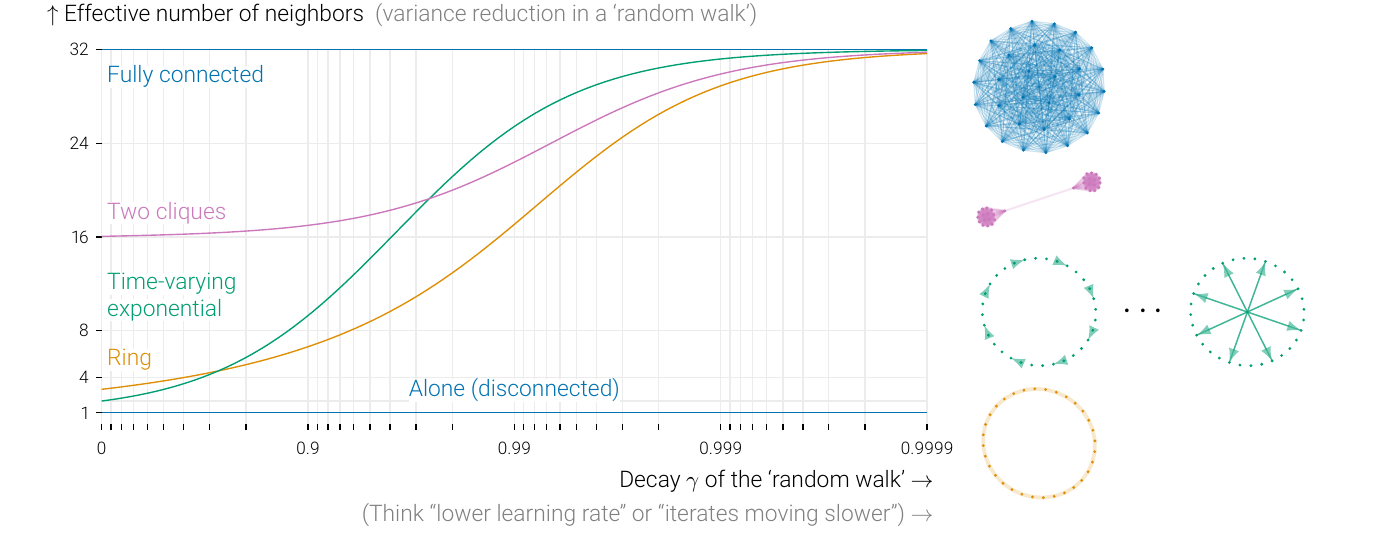}
    \vspace{-17pt}
    \caption{
        \label{fig:eff_nn}
        The effective number of neighbors for several topologies (\autoref{apx:topologies}) measured by their variance reduction in \autoref{eqn:rw}.
        The point $\decay$ on the $x$-axis that matters depends on the learning rate and the task.
        Which topology is `best' varies from problem to problem.
        For large decay rates $\decay$ (corresponding small learning rates), all connected topologies achieve variance reduction close to a fully connected graph.
        For small decay rates (large learning rates), workers only benefit from their direct neighbors (\eg 3 in a ring).
        These curves can be computed explicitly for constant topologies, and simulated efficiently for the time-varying exponential scheme~\cite{assran2019sgp}.
    }
\end{figure}

For regular doubly-stochastic symmetric gossip matrices $\color{tab10_blue}\gossip$ with eigenvalues ${\color{tab10_blue}\weigval}_1, \ldots, {\color{tab10_blue}\weigval}_n$, $\nwrk_{\color{tab10_blue}\gossip}({\decay})$ has a closed-form expression\vspace{-4mm}
\begin{align}
    \nwrk_{\color{tab10_blue}\gossip}({\decay}) = \frac{\frac{1}{1-{\decay}}}{\frac{1}{n}\sum_{i=1}^\nwrk \frac{{\color{tab10_blue}\weigval\idx{i}}^2}{1 - {\color{tab10_blue}\weigval}\idx{i}^2 {\decay}}}. \label{eq:var-reduction}
\end{align}
The notion of variance reduction in random walks, however, naturally extends to infinite topologies or time-varying averaging schemes as well.
\autoref{fig:eff_nn} illustrates $\nwrk_\gossip$ for various topologies.

In our exact characterization of the convergence of \dsgd on the isotropic quadratic toy problem (Appendix~\ref{apx:random-quadratics}),
we find that the effective number of neighbors appears in place of the number of workers $\color{tab10_blue}\nwrk$ in the fully-connected rate of \autoref{eq:centralized}. The rate is the unique solution to
\begin{align} \label{eq:decentralized}
    \rate = 1 - {\color{tab10_red}(1 - \lr)^2} - \frac{\color{tab10_green}(\noise - 1) \lr^2}{\color{tab10_blue} \nwrk_\gossip\big(\frac{(1-\lr)^2}{1 - \rate}\big)}.
\end{align}
For fully-connected and disconnected $\mW$, $\nwrk_{\color{tab10_blue}\gossip}({\decay})=\nwrk$ or 1 respectively, irrespective of $\decay$, and Equation~\ref{eq:decentralized} recovers Equations \ref{eq:solo} and \ref{eq:centralized}.
For other graphs, the effective number of workers depends on the learning rate.
Current theory only considers the case where ${\color{tab10_blue}\nwrk_\gossip} \approx \nwrk$, but the small learning rates this requires can make the term $\color{tab10_red} (1-\lr)^2$ too large, defeating the purpose of collaboration.

Beyond this toy problem, we find that the proposed notion of effective number of neighbors is also meaningful in the analysis of general objectives (\autoref{sec:theory}) and in deep learning (\autoref{sec:experiments}).

%% file: 060_theory.tex
\section{Theoretical analysis}
\label{sec:theory}

In the previous section, we have derived exact rates for a specific function.
Now we present convergence rates for general (strongly) convex functions that are consistent with our observations in the previous section.
We obtain rates that depend on the level of noise, the hardness of the objective, and the topology of the graph.
We will assume the following randomized model for \dsgd:
\begin{equation}\label{eq:def_randomized_dsgd}
	\param\atidx{t+1}{i} = \begin{cases}
		\param\atidx{t}{i} - \lr \nabla f_{\xi\atidx{t}{i}}(\param\atidx{t}{i}) &\text{ with probability } \frac{1}{2}, \\[5pt]
		\sum_{j=1}^\nwrk \gossipweight{ij} \param\atidx{t}{j} &\text{ otherwise,}
	\end{cases}
\end{equation}
where $f_{\xi\atidx{t}{i}}$ represent sampled data points and the gossip weights $\gossipweight{ij}$ are elements of $\gossip$.
This randomized model yields a clean analysis, but similar results hold for standard \dsgd (Appendix~\ref{apx:deterministic-algorithm}).

\begin{assumption}\label{assumption:stochastic}
	The stochastic gradients are such that: (\textsc{i}) $\xi\atidx{t}{i}$ and $\xi\atidx{\ell}{j}$ are independent for all $t, \ell$ and $i \neq j$.
	(\textsc{ii}) $\expect{[ f_{\xi\atidx{t}{i}}]} = f$ for all $t, i$
	(\textsc{iii}) $\expect{\norm{\nabla f_{\xi\atidx{t}{i}}(\param\at{\star})}^2} \leq \variance^2$ for all $t, i$, where $\param\at{\star}$ is a minimizer of $f$.
	(\textsc{iv}) $f_{\xi\atidx{t}{i}}$ is  convex and $\noise$-smooth for all $t,i$.
	(\textsc{v}) $f$ is $\mu$-strongly-convex for $\mu\ge0$ and $\smoothf$-smooth.
\end{assumption}

The smoothness $\noise$ of the stochastic functions $f_{\xi}$ defines the level of noise in the problem (the lower, the better).
The ratio $\noise / \smoothf$ compares the difficulty of optimizing with stochastic gradients to the difficulty with the true global gradient (before reaching the `variance region' of distance $\cO(\variance^2)$ to the optimum).
Assuming better smoothness for the global average objective than for the local functions is key to showing the benefit of averaging between workers.
Without communication, convergence to the variance region is ensured for learning rates $\lr \leq 1/\noise$.
If $\noise \approx \smoothf$, there is little noise and cooperation does not help before $\norm{\param\at{t} - \param\at{\star}}^2 \approx \variance^2$.
Yet, in noisy regimes ($\noise \ggg \smoothf$), such as in \autoref{sec:toy-model} in which $\noise = \dims+2 \ggg 1 = \smoothf$, averaging enables larger step-sizes up to $\min(1/\smoothf, \nwrk/\noise)$, greatly speeding up the initial training phase.
This is precisely what we prove in Theorem~\ref{thm:convex_rates}.

\emph{If} the workers always remain close ($\param\idx{i} \approx \frac{1}{n} (\param\idx{1} + \ldots + \param\idx{n})\;\forall i$, or equivalently $\frac{1}{n} \bm{1}\bm{1}^\top \param \approx \param$), \dsgd behaves the same as SGD on the average parameter $\frac{1}{n}\sum_{i=1}^\nwrk \param\idx{i}$, and the learning rate depends on $\max(\noise / \nwrk, \smoothf)$, showing a reduction of variance by $\nwrk$.
To maintain ``$\frac{1}{n}\bm{1}\bm{1}^\top \param \approx \param$'', however, we require a small learning rate.
This is a common starting point for the analysis of \dsgd, in particular for the proofs in \citet{koloskova2020unified}.
On the other extreme, if we do not assume closeness between workers, ``$\mI \mspace{1mu} \param \approx \param$'' always holds.
In this case, there is no variance reduction, but no requirement for a small learning rate either.
In \autoref{sec:toy-model}, we found that, at the optimal learning rate, workers are \emph{not} close to all other workers, but they \emph{are} close to others that are not too far away in the graph.

We capture the concept of `local closeness' by defining an averaging matrix $\mmat$.
It allows us to consider semi-local averaging beyond direct neighbors, but without fully averaging with the whole graph.
We ensure that ``$\mmat \param \approx \param$'', leading to some improvement in the smoothness between $\noise$ and $\noise / \nwrk$, interpolating between the two previous cases.
Each matrix $\mmat$ implies a requirement on the learning rate, as well as an improvement in smoothness. Based on \autoref{sec:toy-model}, we therefore focus on a specific family of matrices that strike a good balance between the two:
We choose $\mmat$ as the covariance of a decay-$\decay$ `random walk process' with the graph, meaning that
\begin{equation}\label{eq:M_definition}
\mmat  = (1 - \decay)\sum_{k=1}^\infty \decay^{k - 1} \gossip^{2k} = (1-\decay)\gossip^2 (1- \decay \gossip^2)^{-1}.
\end{equation}
Varying $\decay$ induces a spectrum of averaging neighborhoods from $\mmat = \gossip^2$ ($\decay = 0$) to $\mmat = \frac{1}{n}\bm{1}\bm{1}^\top$ ($\decay = 1$).
$\decay$ also implies an effective number of neighbors $\nwrk_\gossip(\decay)$: the larger $\decay$, the larger $\nwrk_\gossip(\decay)$.

Theorem~\ref{thm:convex_rates} provides convergence rates for any value of $\decay$, but the best rates are obtained for a specific~$\decay$ that balances the benefit of averaging with the constraint it imposes on closeness between neighbors.
In the following theorem, we assume that $\mweight{ii} = \mweight{jj}$ for all $i,j$, so that $\mweight{ii}^{-1} = \neff$: the effective number of neighbors defined in~\eqref{eq:var-reduction} is equal to the inverse of the self-weights of matrix $\mmat$.
Otherwise, all results hold by replacing $\neff$ with $\min_i \mweight{ii}^{-1}$.
\vspace{6pt plus2pt minus2pt}
\begin{theorem}\label{thm:convex_rates}
	If Assumption~\ref{assumption:stochastic} holds, and the learning rate satisfies
	\Hadrien{We can also just use $\beta \geq \frac{1 - \decay}{2}$}
	\begin{equation}\label{eq:lr_conditions}
		\lr \leq \min\left( \frac{1}{8(\noise / \neff + \smoothf)},  \frac{1 - \decay \lmax(\gossip)}{2\neff \smoothf}\right),
	\end{equation}
	then the iterates obtained by~\eqref{eq:def_randomized_dsgd} verify
	\begin{equation}
		\norm{\param\at{t} - \param\at{\star}}^2_\mmat + \frac{1}{\neff} \norm{\param\at{t}}^2_{\mI - \mmat} \leq \left(1 - \frac{\lr \mu}{2}\right)^t C_0 + \frac{8 \lr \variance^2}{\neff},
	\end{equation}
\end{theorem}

The bound on the learning rate~\eqref{eq:lr_conditions} represents the tension between (\textsc{i}) reducing the noise $\noise$ by averaging with more people (larger $\neff$), which is the first term in the minimum, and (\textsc{ii}) staying close to all of them.
A large spectral gap $1 - \lmax(\gossip)$ reduces the second constraint, but we allow non-trivial learning rates $\lr > 0$ even when $\lmax(\gossip) = 1$ (infinite graphs) as long as $\decay < 1$.

\autoref{thm:convex_rates} gives a rate for each parameter $\decay$ that controls the local neighborhood size.
The task that remains is to find the $\decay$ parameter that gives the best convergence guarantees (the largest learning rate).
As explained before, one should never reduce the learning rate in order to be close to others, because the goal of collaboration is to \emph{increase} the learning rate.
We should therefore pick $\decay$ such that the first term in Equation~\eqref{eq:lr_conditions} dominates.
This intuition is summarized in Corollary~\ref{corr:rates}, which compares the performance of \dsgd with centralized SGD with fewer workers.

\vspace{6pt plus2pt minus2pt}
\begin{corollary}\label{corr:rates}
	\dsgd is as fast as centralized mini-batch SGD with $\cO(\neff)$ workers, assuming that $\noise \geq \nwrk \smoothf$, and that the parameter $\decay$ is the highest $\decay$ such that $\frac{2\neff^2}{1 - \decay \lmax(\gossip)} \leq 32\frac{\noise}{\smoothf}$.
	This corresponds to a learning rate $\lr = \neff / 16\noise$. \vspace{5pt}
\end{corollary}

\begin{figure}
	\includegraphics{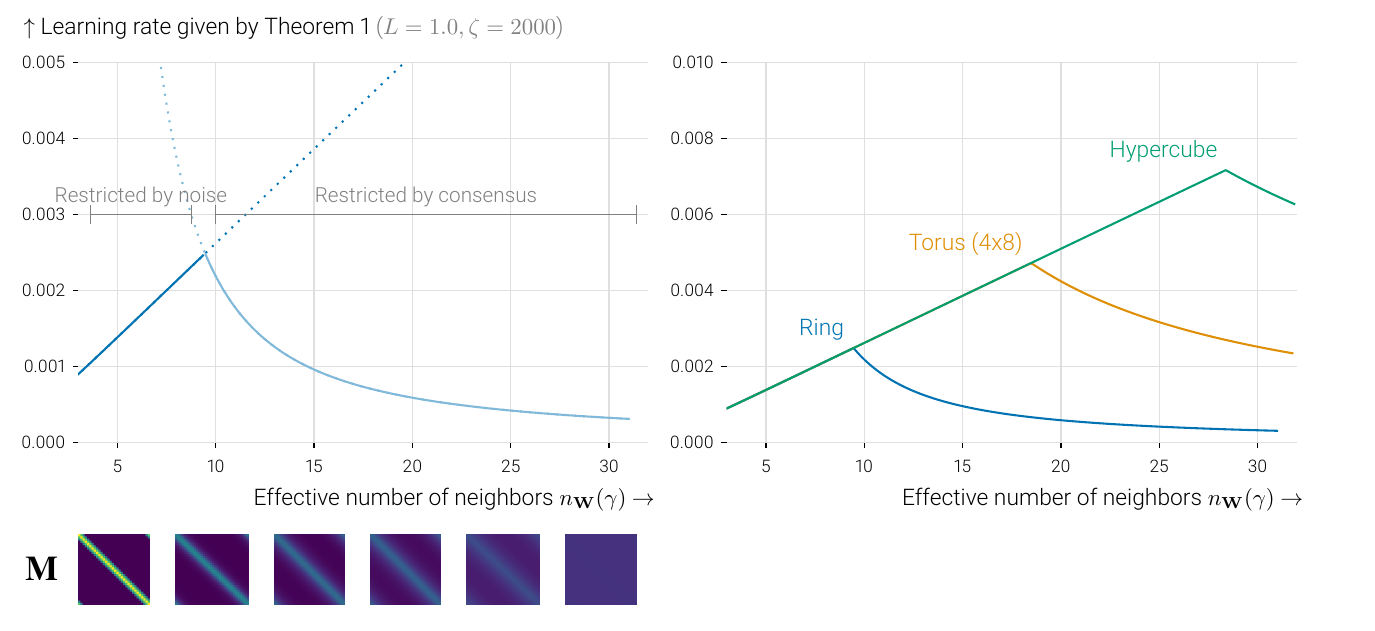}
    \centering
	\vspace{-20pt}
    \caption{
		\label{fig:lr_gamma_theory}
		Maximum learning rates prescribed by \autoref{thm:convex_rates}, varying the parameter $\decay$ that implies an effective neighborhood size ($x$-axis) and an averaging matrix $\mM$ (drawn as \emph{heatmaps}).
		On the \emph{left}, we show the details for a 32-worker ring topology, and on the \emph{right}, we compare it to more connected topologies.
	 	Increasing $\decay$ (and with it $\nwrk_\gossip(\decay)$) initially leads to larger learning rates thanks to noise reduction.
		At the optimum, the cost of consensus exceeds the benefit of further reduced noise.
	}
\end{figure}

The typical \dsgd learning rates~\cite{koloskova2020unified} are of order $\cO(\min(1/T, 1-\lambda_2(W)))$, which are much smaller than the learning rate of Corollary~\ref{corr:rates} when $\lambda_2(\gossip)$ is large or the number of iterations large. We use the condition $\noise \geq \nwrk \smoothf$ only to present results in a simpler way.
The condition $\frac{2\neff^2}{1 - \decay \lmax(\gossip)}$ only depends on the size and topology of the graph, and can easily be computed in many cases.
Thus, to obtain the best guarantees, we start from $\decay = 0$ and then increase it until either $\neff \approx \nwrk$, the total size of the graph, or the two terms in the minimum match.
This is how we obtain \autoref{fig:lr_gamma_theory}.

\vspace{-3mm}
\begin{proof}[Proof sketch (Theorem~\ref{thm:convex_rates})]
	The proof relies on a simple argument: rather than bounding $\norm{\param\at{t} - \param\at{\star}}^2$ or $\norm{\frac{1}{n}\bm{1}\bm{1}^\top  \param\at{t} - \param\at{\star}}^2$, we analyze $\norm{\param\at{t} - \param\at{\star}}^2_\mmat$.
	This term better captures the benefit of averaging than $\norm{\param\at{t} - \param\at{\star}}^2$, thus leading to better smoothness constants, as long as $\norm{\param\at{t}}^2_{\mI - \mmat}$ is not too large.
	This yields fast rates without the need to guarantee that iterates between very distant workers remain close, which would be prohibitively expensive.
\end{proof}
\vspace{-2mm}

Theorem~\ref{thm:convex_rates} is a special case of a more general theorem presented in Appendix~\ref{apx:theory-convex}.
This version, among other things, covers different choices of parameters, unbalanced communication and computation probabilities (thus allowing for local steps), and the convex ($\mu = 0$) case.

%% file: 065_experiments.tex
\section{Experimental analysis}
\label{sec:experiments}

While in the previous sections we have discussed isotropic quadratics or convex and smooth functions, the initial motivation for this work comes from observations in deep learning.
First, it is crucial in deep learning to use a large learning rate in the initial phase of training~\cite{li2019largelr}.
Contrary to what current theory prescribes, we do not use smaller learning rates in decentralized optimization than when training alone (even when data is heterogeneous.)
And second, we find that the spectral gap of a topology is not predictive of the performance of that topology in deep learning experiments.

In this section, we experiment with a variety of 32-worker topologies on Cifar-10~\cite{cifar10} with a VGG-11 model~\cite{simonyan2015vgg}.
Like other recent works~\cite{lin2021quasiglobal,vogels2021relay}, we opt for this older model, because it does not include BatchNorm~\cite{ioffe2015batchnorm} which forms an orthogonal challenge for decentralized SGD.
Please refer to \autoref{apx:cifar-details} for full details on the experimental setup.
Our set of topologies (\autoref{apx:topologies}) includes regular graphs like rings and toruses, but also irregular graphs such as a binary tree~\cite{vogels2021relay} and social network~\cite{davis1930socialwomen}, and a time-varying exponential scheme~\cite{assran2019sgp}.
We focus on the initial phase of training, 25k steps in our case, where both train and test loss converge close to linearly.
Using a large learning rate in this phase is found to be important for good generalization~\cite{li2019largelr}.

\newcommand{\kilo}{\hspace{1pt}k\xspace}

\begin{figure}
    \includegraphics{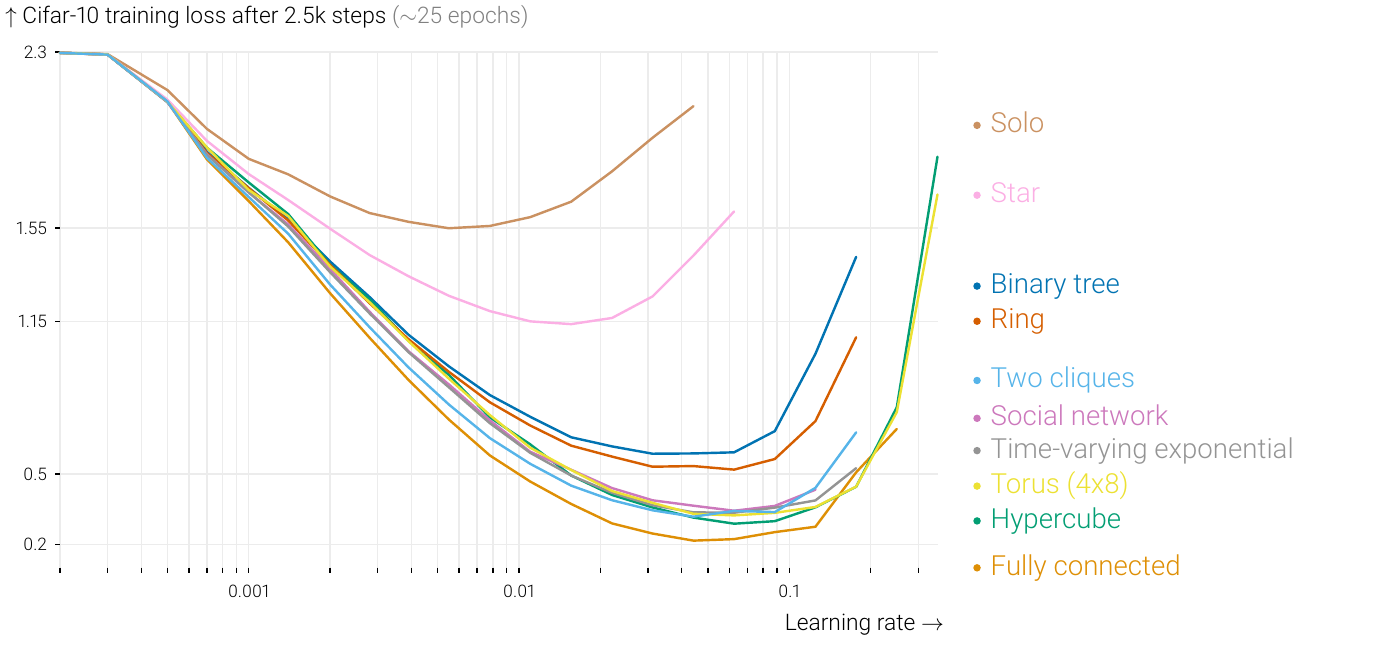}
    \centering
    \vspace{-18pt}
    \caption{
        \label{fig:results-cifar}
        Training loss reached after 2.5\kilo SGD steps with a variety of graph topologies.
        In all cases, averaging yields a small increase in speed for small learning rates, but a large gain over training alone comes from being able to increase the learning rate.
        While the star has a better spectral gap (0.031) than the ring (0.013), it performs worse, and does not allow large learning rates.
        For reference, similar curves for fully-connected graphs of varying sizes are in \autoref{apx:experiments}.
    }
\end{figure}

\autoref{fig:results-cifar} shows the loss reached after the first 2.5\kilo SGD steps for all topologies and for a dense grid of learning rates.
The curves have the same global structure as those for isotropic quadratics \autoref{fig:intro_illustration}: (sparse) averaging yields a small increase in speed for small learning rates, but a large gain over training alone comes from being able to increase the learning rate.
The best schemes support almost the same learning rate as 32 fully-connected workers, and get close in performance.

\begin{figure}
    \includegraphics[width=0.99\linewidth]{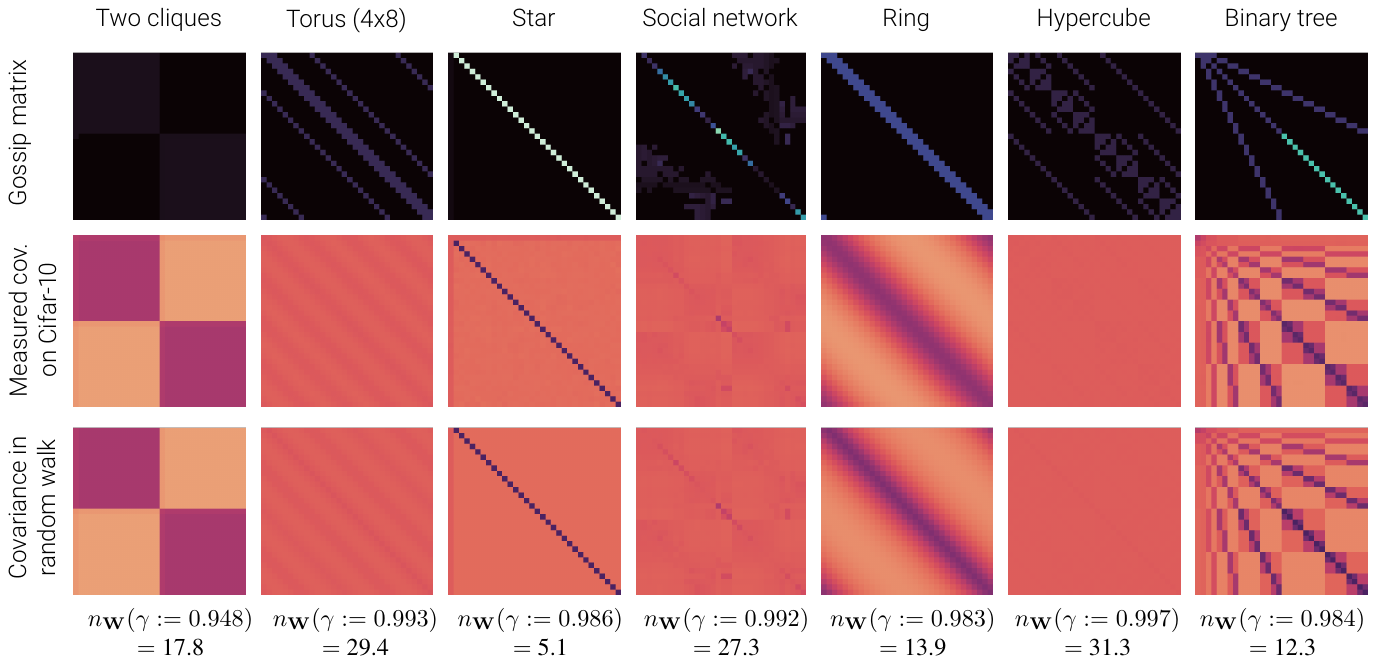}
    \vspace{-10pt}
    \centering
    \caption{
        \label{fig:cifar-covariance}
        Measured covariance in Cifar-10 (second row) between workers using various graphs (top row).
        After 10 epochs, we store a checkpoint of the model and train repeatedly for 100 SGD steps, yielding 100 models for 32 workers.
        We show normalized covariance matrices between the workers.
        These are very well approximated by the covariance in the random walk process of \autoref{sec:toy-model} (third row).
        We print the fitted decay parameters and corresponding `effective number of neighbors'.
        \vspace{-10pt}
    }
\end{figure}

We also find that the random walks introduced in \autoref{sec:toy-model} are a good model for variance between workers in deep learning.
\autoref{fig:cifar-covariance} shows the empirical covariance between the workers after 100 SGD steps.
Just like for isotropic quadratics, the covariance is accurately modeled by the covariance in the random walk process for a certain decay rate $\decay$.

\begin{figure}
    \includegraphics{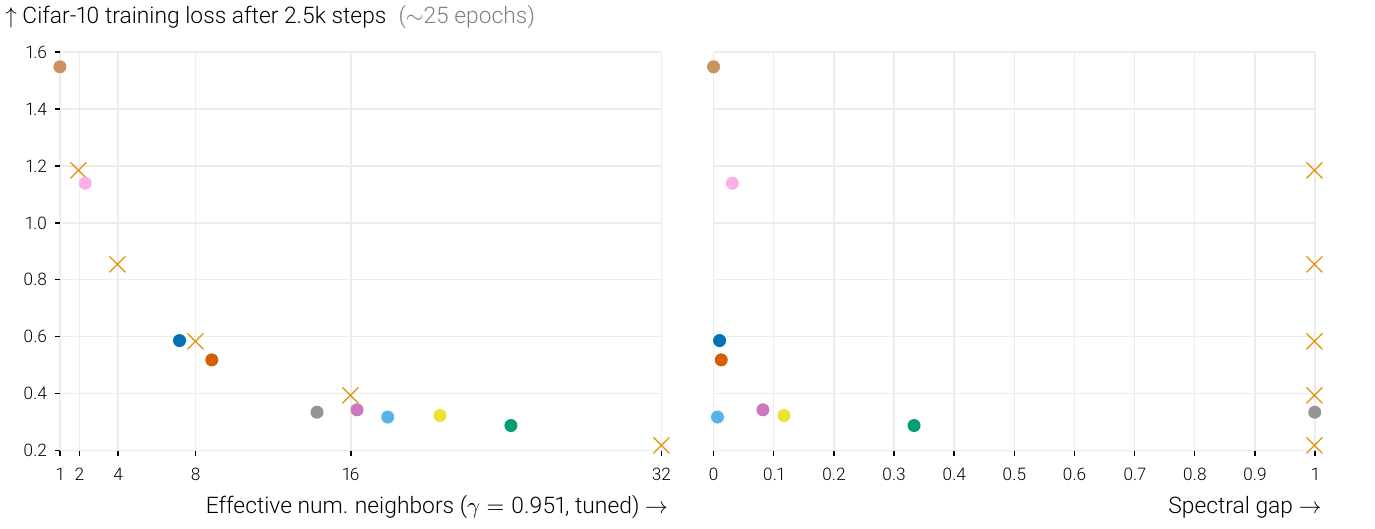}
    \centering
    \caption{
        Cifar-10 training loss after 2.5\kilo steps
        for all studied topologies with their optimal learning rates.
        Colors match \autoref{fig:results-cifar}, and {\color{tab10_orange}$\times$} indicates fully-connected graphs with varying number of workers.
        After fitting a decay parameter $\decay=0.951$ that captures problem specifics, the effective number of neighbors (left) as measured by variance reduction in a random walk (like in \autoref{sec:toy-model}) explains the relative performance of these graphs much better than the spectral gap of these topologies (right).
        \label{fig:correlation_plots}
    }
\end{figure}

Finally, we observe that the effective number of neighbors computed by the variance reduction in a random walk (\autoref{sec:toy-model}) accurately describes the relative performance under tuned learning rates of graph topologies on our task, including for irregular and time-varying topologies.
This is in contrast to the topology's spectral gaps, which we find to be not predictive.
We fit a decay rate $\decay=0.951$ that seems to capture the specifics of our problem, and show the correlation in \autoref{fig:correlation_plots}.

In Appendix~\ref{apx:fashion}, we replicate the same experiments in a different setting.
There, we use larger graphs (of 64 workers), a different model and data set (an MLP on Fashion MNIST~\cite{xiao2017fashion}), and no momentum or weight decay.
The results in this setting are qualitatively comparable to the ones presented above.

%% file: 070_conclusion.tex
\section{Conclusion}

We have shown that the sparse averaging in decentralized learning allows larger learning rates to be used, and that it speeds up training.
With the optimal large learning rate, the workers' models are not guaranteed to remain close to their global average.
Enforcing global consensus is unnecessary in the \iid setting and the small learning rates it would require are counter-productive.
With the optimal learning rate, models \emph{do} remain close to some local average in a weighted neighborhood around them.
The workers benefit from a number of `effective neighbors', smaller than the whole graph, that allow them to use a large learning rate while retaining sufficient consensus within the `local neighborhood'.

Based on our insights, we encourage practitioners of sparse distributed learning to look beyond the spectral gap of graph topologies, and to investigate the actual `effective number of neighbors' that is used.
We also hope that our insights motivate theoreticians to be mindful of assumptions that artificially limit the learning rate. \Thijs{Up for debate $\ldots$. The point of this paragraph was to answer the question `what can this be used for?'}

We show experimentally that our conclusions hold in deep learning, but extending our theory to the non-convex setting is an important open direction that could reveal interesting new phenomena.
Furthermore, an extension of our semi-local analysis to the heterogeneous setting where workers optimize different objectives could shed further light on the practical performance of \dsgd.

\break

%% file: 080_acknowledgements.tex
\begin{ack}
	This project was supported by SNSF grant 200020\_200342.

	We thank Lie He for valuable conversations and for identifying the discrepancy between a topology's spectral gap and its empirical performance.
	We also thank Rapha\"el Berthier, Aditya Vardhan Varre and Yatin Dandi for their feedback on the manuscript.
\end{ack}

%% file: 090_checklist.tex
\section*{Checklist}

\begin{enumerate}

  \item For all authors...
        \begin{enumerate}
          \item Do the main claims made in the abstract and introduction accurately reflect the paper's contributions and scope?
                \answerYes{}
          \item Did you describe the limitations of your work?
                \answerYes{}
          \item Did you discuss any potential negative societal impacts of your work?
                \answerYes{}
          \item Have you read the ethics review guidelines and ensured that your paper conforms to them?
                \answerYes{}
        \end{enumerate}

  \item If you are including theoretical results...
        \begin{enumerate}
          \item Did you state the full set of assumptions of all theoretical results?
                \answerYes{}
          \item Did you include complete proofs of all theoretical results?
                \answerYes{}
        \end{enumerate}

  \item If you ran experiments...
        \begin{enumerate}
          \item Did you include the code, data, and instructions needed to reproduce the main experimental results (either in the supplemental material or as a URL)?
                \answerYes{}
          \item Did you specify all the training details (e.g., data splits, hyperparameters, how they were chosen)?
                \answerNA{}
          \item Did you report error bars (e.g., with respect to the random seed after running experiments multiple times)?
                \answerYes{}
          \item Did you include the total amount of compute and the type of resources used (e.g., type of GPUs, internal cluster, or cloud provider)?
                \answerNo{}
        \end{enumerate}

  \item If you are using existing assets (e.g., code, data, models) or curating/releasing new assets...
        \begin{enumerate}
          \item If your work uses existing assets, did you cite the creators?
                \answerYes{}
          \item Did you mention the license of the assets?
                \answerNo{}
          \item Did you include any new assets either in the supplemental material or as a URL?
                \answerNA{}
          \item Did you discuss whether and how consent was obtained from people whose data you're using/curating?
                \answerNA{}
          \item Did you discuss whether the data you are using/curating contains personally identifiable information or offensive content?
                \answerNA{}
        \end{enumerate}

  \item If you used crowdsourcing or conducted research with human subjects...
        \begin{enumerate}
          \item Did you include the full text of instructions given to participants and screenshots, if applicable?
                \answerNA{}
          \item Did you describe any potential participant risks, with links to Institutional Review Board (IRB) approvals, if applicable?
                \answerNA{}
          \item Did you include the estimated hourly wage paid to participants and the total amount spent on participant compensation?
                \answerNA{}
        \end{enumerate}

\end{enumerate}

%% file: 095_appendix.tex
\appendix

{
    \hypersetup{hidelinks}
    \parskip=0em
    \renewcommand{\contentsname}{Contents of the Appendix}
    \tableofcontents
    \addtocontents{toc}{\protect\setcounter{tocdepth}{3}}
    \vfill
    \break
}

\section{Notation}
\label{apx:notation}
\autoref{apx:tab:notation} defines some notation and conventions used throughout this paper and in the appendix.%
\begin{table}[h]
    \centering
    \caption{Notation\label{apx:tab:notation}}
    \begin{tabularx}{.55\textwidth}{ll}
        \toprule
        Bold symbol $\vv$  & Vector \\
        Bold uppercase $\mM$  & Matrix \\
        $\cN^d(0, 1)$  & Standard normal distribution \\
                       & with $d$ independent dimensions \\
        $\lin{\xx, \yy}$ & Inner product $\xx^\top \yy$ \\
        $\norm{\mT}_2$ & Spectral norm \\ %
        $\norm{\mT}_F$ & Frobenius norm \\ %
        $\mP \otimes \mQ$ & Kronecker product \\
        $\bm{1}$ & Vector of all ones \\
        \bottomrule
    \end{tabularx}
\end{table}%

\section{Topologies}
\label{apx:topologies}

\begin{figure}[h]
    \includegraphics[scale=0.97]{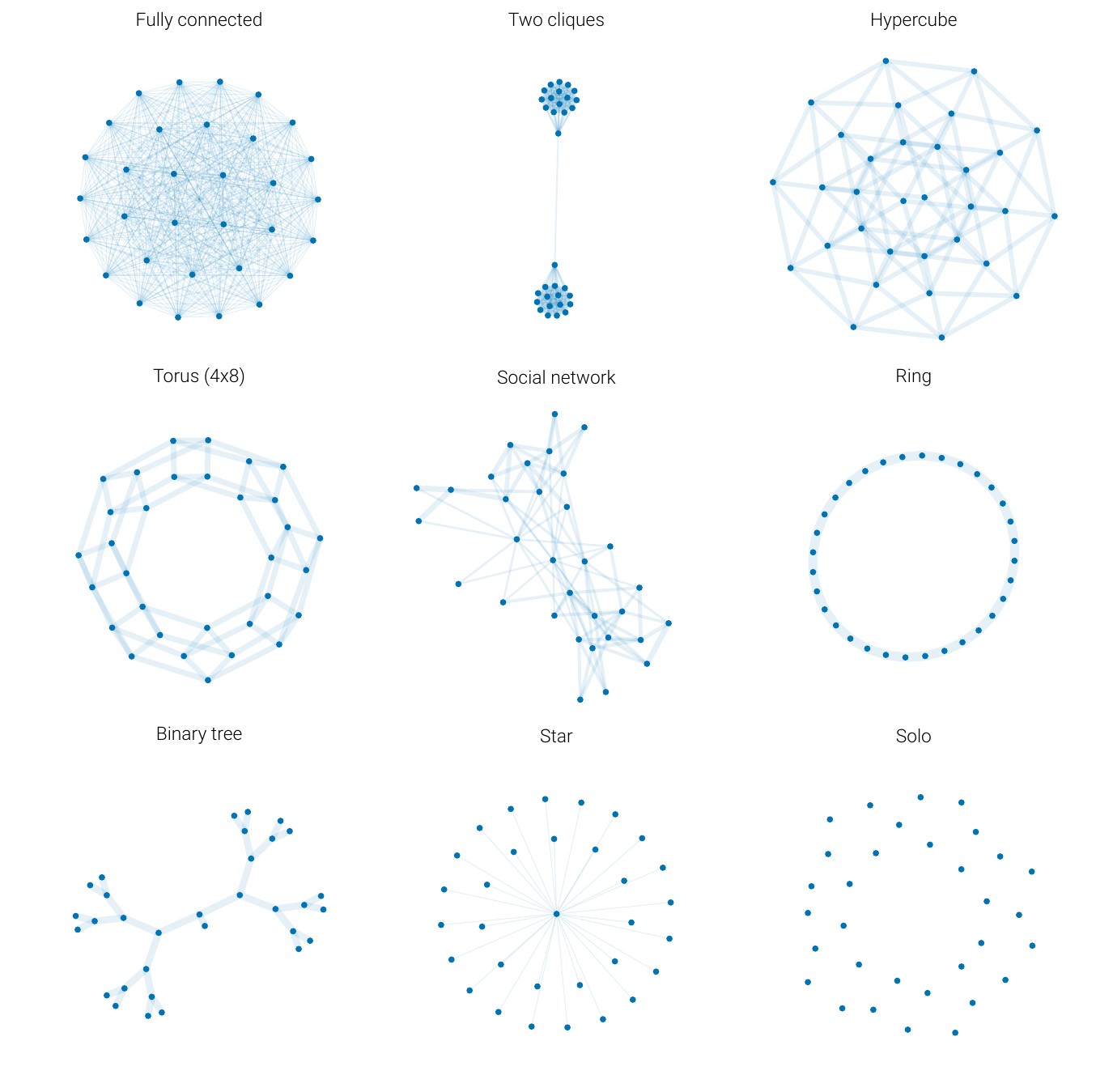}\vspace{-6mm}
    \caption{
        \label{apx:fig:graph-topos}
        Spring-layout drawings of the static graph topologies considered used this paper.
        The nodes represent workers, and an edge between two workers indicates that they are connected.
        The thickness of a edges is proportional to its averaging weight.
    }
\end{figure}

\begin{figure}[h]
    \includegraphics{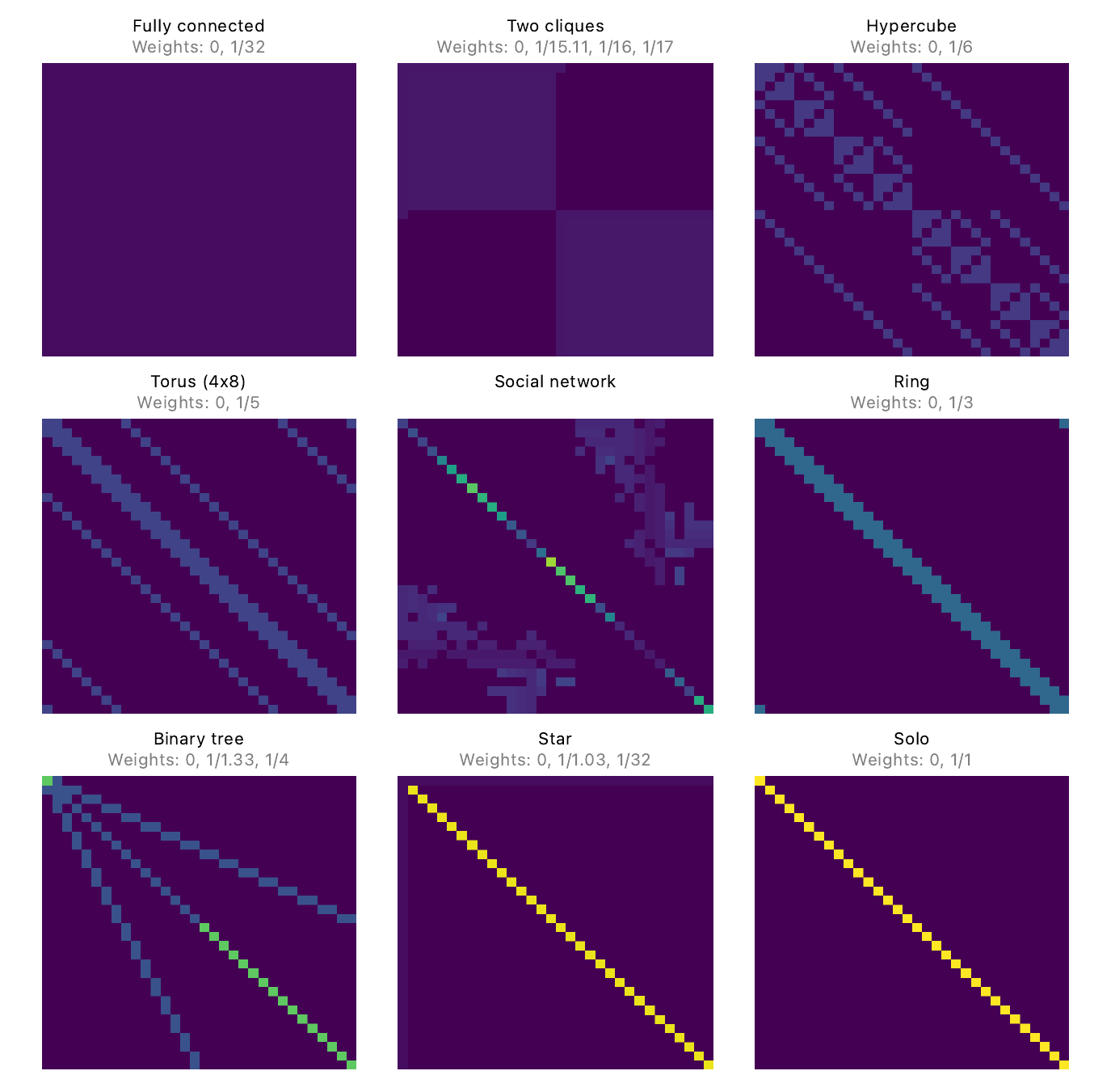}\vspace{-3.2mm}
    \caption{
        \label{apx:fig:gossip-matrices}
        Gossip matrices corresponding to the graph topologies drawn in \autoref{apx:fig:graph-topos}.
        $x$ and $y$ axes represent workers,
        and the color of each coordinate in the plots indicates the gossip weight between each pair of workers.
        The brigher, the higher the weight.
    }
\end{figure}

\begin{figure}[h]
    \includegraphics{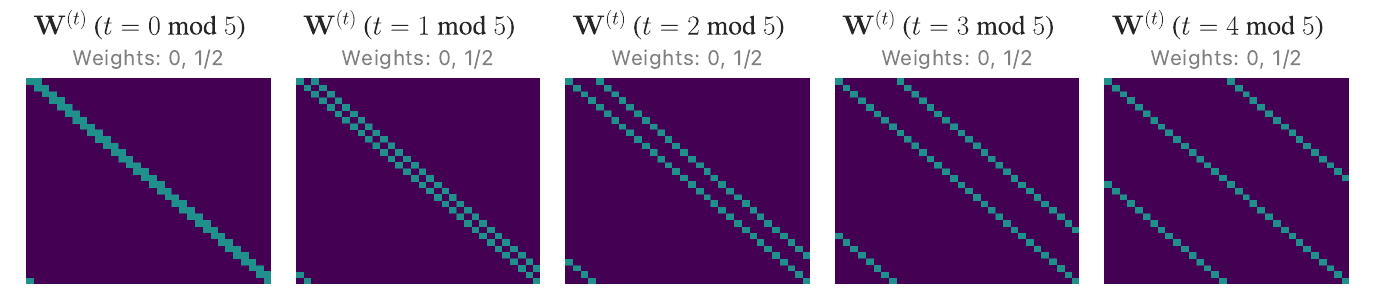}\vspace{-3.2mm}
    \caption{
        \label{apx:fig:time-varying-gossip}
        Gossip matrices for the time-varying exponential graph~\cite{assran2019sgp,ying2021exponential}.
        The product of $\log \nwrk$ consecutive gossip matrices equals to the fully-connected averaging matrix with $\gossipweight{ij}=1/\nwrk\;  \forall i,j$.
    }
\end{figure}

The static topologies that we consider in this work are drawn in \autoref{apx:fig:graph-topos}.
Figures \ref{apx:fig:gossip-matrices} and \ref{apx:fig:time-varying-gossip} show the gossip matrices we use in detail.

\input{095_appendix_10_random_quadratics}

\input{095_appendix_20_theory}

\input{095_appendix_70_cifar_details}

\input{095_appendix_30_experiments}

%% file: 095_appendix_10_random_quadratics.tex
\clearpage
\section{Random quadratics}
\label{apx:random-quadratics}

\subsection{Objective}

We study the simple problem of minimizing an isotropic $\dims$-dimensional quadratic,
$$
    \param\at{\star} = \argmin_{\param \in \R^d} f(\param)
$$
where the objective function $f(x) = \frac{1}{2}\norm{\param}^2$ is considered to be the expectation over an infinite dataset with random normal features and labels 0:
\begin{align}
f(\param) = \expect_{\randdata \sim \cN^\dims(0,1)} \frac{1}{2} \lin{\randdata, \param}^2.
\label{eq:objective}
\end{align}
The optimum of this objective is at $\param\at{\star} = \0$ without loss of generality,
because any shifted quadratic would behave the same in the algorithm studied.
We will access this objective function through stochastic gradients of the form $\gg(\param) = \randdata \randdata^\top \param$.
The stochasticity of these gradients disappears at the optimum, like in an over-parameterized model.

The difficulty of this problem depends on the dimensionality $\dims$.
For a lower-dimensional problem, the `stochastic Hessian' $\randdata \randdata^\top$ is closer to the true hessian $\mI$ than for a high dimensional one.
This level of stochasticity is captured by the following quantity:
\begin{definition}[Noise level]
    $\noise = \sup_\param \frac{\expect_\randdata \norm{\randdata \randdata^\top \param}^2}{\norm{\param}^2}$.
\end{definition}

For our random normal data with batch size 1, this notion of noise level corresponds directly to the dimensionality of the data as $\noise = \dims + 2$.

\subsection{Algorithm}

The objective \eqref{eq:objective} is collaboratively optimized by $n$ workers.
At every time step $t$, each worker $i$ has its own copy of the `model' $\param\atidx{t}{i} \in \R^\dims$.
In the \dsgd algorithm, workers iteratively compute stochastic gradient estimates $\gg\atidx{t}{i} = \randdata\atidx{t}{i} \randdata{\atidx{t}{i}}^\top \param\atidx{t}{i}$, where $\randdata\atidx{t}{i}$ are \iid from $\cN^\dims(0, 1)$.
The stochastic gradients are unbiased: $\expect{\gg\atidx{t}{i}} = \nabla f(\param\atidx{t}{i}) = \param\atidx{t}{i}$.

Workers interleave stochastic gradient updates with gossip averaging:
\begin{align*}
    \param\atidx{0}{i} &= \param\at{0} \quad \forall i \\
    \param\atidx{t+1}{i} &= \gossip(\param\atidx{t}{i} - \lr \mspace{1mu}  \gg\atidx{t}{i}),
\end{align*}
where $\lr$ is the learning rate and
\begin{align*}
    \gossip(\param\idx{i}) = \sum_{j=1}^\nwrk \gossipweight{ij} \param\idx{j}.
\end{align*}
This linear operation can be interpreted as matrix multiplication, but one operating on each coordinate of the model independently.
$\gossip$ is an $\nwrk \times \nwrk$ matrix, and \emph{not} a $\dims \times \dims$ matrix as the notation may suggest.
The averaging weights $\gossipweight{ij}$ encode the connectivity of the communication topology: non-zero $\gossipweight{ij}$ implies that workers $i$ and $j$ are directly connected.
We make several assumptions about the gossip weights in this analysis:

\begin{assumption}
    \label{assumption:constant}
    Constant gossip weights: The weights $\gossipweight{ij}$ do not change between steps of \dsgd.
\end{assumption}

\begin{assumption}
    \label{assumption:symmetric}
    Symmetric gossip weights: $\gossipweight{ij} = \gossipweight{ji}$.
\end{assumption}

\begin{assumption}
    \label{assumption:doubly_stochastic}
    Doubly stochastic gossip weights: $\gossipweight{ij} \geq 0 \; \forall i, j$, $\sum_j \gossipweight{ij} = 1 \; \forall i$, $\sum_i \gossipweight{ij} = 1\;\forall j$.
\end{assumption}

\begin{assumption}
    \label{assumption:regular}
    Regular topology: all workers have $k$ directly-connected neighbors, and $\gossipweight{ij} = c$ for some constant $c$, and for each edge where $i\neq j$.
\end{assumption}

\begin{definition}
    Spectrum of $\gossip$.
    Let the eigenvalues of $\gossip$ be $\weigval_1 \geq \weigval_2 \geq \ldots \geq \weigval_n$.
    We call the corresponding eigenvectors $\weigvec_1, \ldots, \weigvec_n$.
    Under assumption \ref{assumption:doubly_stochastic}, $\weigval_1=1$, and we call $1 - \weigval_2$ the \emph{spectral gap} of $\gossip$.
\end{definition}

The assumptions on constant gossip weights and regular topologies are mainly here to ease the analysis.
We experimentally observe that our findings hold for time-varying topologies and infinite graphs, too, and that they approximately hold for irregular graphs.

\subsection{Linear convergence of an unrolled error vector}

We will study the convergence of the algorithm by tracking the \emph{error matrix} $\emat \in \R^{\nwrk \times \nwrk}$.
The coordinates of this matrix are the expected covariance between each pair of workers in the network.
$$
    \emat\atidx{t}{ij} = \expect \lin{\param\atidx{t}{i}, \param\atidx{t}{j}}.
$$

We sometimes flatten the error matrix into a vector $\evec \in \R^{\nwrk^2}$, such that $\evec_{\nwrk i+j} = \emat_{ij}$.
The diagonal entries of this matrix describe the worker's error on the objective, and as all workers converge to the optimum at zero, each entry of the matrix will converge to zero.
Our analysis of $\emat$ quantity starts with a key observation:

\vspace{2mm}
\begin{lemma}
    There exists an $\nwrk^2 \times \nwrk^2$ `transition' matrix $\trans$ such that $\evec\at{t+1} = \trans \mspace{0.5mu} \evec\at{t} \; \forall t$.
\end{lemma}
\begin{proof}
    Because both gossip averaging and the gradient updates are linear, this follows from expanding the inner product.
\end{proof}

The transition matrix $\trans$ depends on the gossip matrix $\gossip$ and on the learning rate $\lr$.
Its spectral gap describes the convergence of the algorithm.
\dsgd converges linearly if the norm $\norm{\trans}_2 < 1$.

We separate $\trans$ into a product $\trans=\trans^\text{gossip} \trans^\text{grad}$, where $\trans^\text{grad}$ and $\trans^\text{gossip}$ respectively capture the gradient update and gossip steps of the algorithm.
We find that
\begin{align*}
    \trans^\text{gossip} = \gossip \otimes \gossip
\end{align*}
and that $\trans^\text{grad}$ is diagonal.
It only operates element-wise, such that
\begin{align}
    \label{eqn:cov-sgd}
    \left[\trans^\text{grad} \evec \right]_{\nwrk i + j} = \begin{cases}
        (1-\lr)^2 \evec_{\nwrk i + j} + (\noise - 1) \lr^2 \evec_{\nwrk i + j} & i = j \text{ (same worker)}, \\
        (1-\lr)^2 \evec_{\nwrk i + j} & i \neq j \text{ (different workers)}.
    \end{cases}
\end{align}
This follows directly from expanding the inner product $\lin{\param\idx{i} - \lr \gg_i, \param\idx{j} - \lr \gg_j}$.
The terms with $i=j$ behave differently than the ones where $i \neq j$, because the noise cancels if $i\neq j$.

\subsection{Random walks with gossip averaging}

Before we study the convergence of \dsgd on the random quadratic objective, we first take a step back and inspect a particular random walk process, where workers average their random walk iterates through gossip averaging.

Let $\rw\at{t} \in \R^\nwrk$ be a vector containing (scalar) iterates of $\nwrk$ workers in the following process:
\begin{align}
    \rw\at{0} &= \0, \\
    \rw\at{t+1} &= \gossip\left(\sqrt{\decay} \rw\at{t} +\rwn\at{t}\right) \quad \text{ where } \rwn\at{t} \sim \cN^\nwrk(0, 1).
\end{align}
We call the parameter $0<\decay\leq1$ the `decay rate'.
Note that the name \emph{random walk} refers to iterative addition of random noise to the workers iterates, and not to a `random walk' between nodes of the graph.

For this random walk, we will track the covariance matrix $\rwcov \in \R^{\nwrk \times \nwrk}$ across workers (and its flattened version $\rwcovv \in \R^{\nwrk^2}$).
Its coordinates are
\begin{align*}
    \rwcov\atidx{t}{ij} = \expect [\rw\atidx{t}{i} \rw\atidx{t}{j}].
\end{align*}

\vspace{2mm}
\begin{lemma}
    \label{lemma:rwalk-cov}
    For static, symmetric and doubly-stochastic topologies (Assumptions~\ref{assumption:constant}, \ref{assumption:symmetric} and \ref{assumption:doubly_stochastic}), the Eigen decomposition of the covariance is
    \begin{align*}
    \rwcov\at{t} = \sum_{i=1}^\nwrk \rwcoveig\atidx{t}{i} \weigvec_i \weigvec_i^\top,
    \end{align*}
    with $0 \leq \rwcoveig\atidx{t}{i} \leq \frac{\weigval_i^2}{1 - \decay \weigval_i^2}$.
    Here $(\weigval_i, \weigvec_i)$ are the eigenvalue/eigenvector pairs of $\gossip$.
    As $t \to \infty$, $\rwcoveig\atidx{t}{i} = \frac{\weigval_i^2}{1 - \decay \weigval_i^2}$ with equality.
\end{lemma}
\begin{proof}
    We can unroll the iterations:
    \begin{align*}
        \rw\at{t} = \sum_{k=1}^{t}\gossip^{k} \decay^{(k-1)/2} \rwn\at{t-k}
    \end{align*}
    and use the temporal independence of $\rwn\at{t}$ to write
    \begin{align*}
        \rwcov\at{t} = \expect [ \rw\at{t} {\rw\at{t}}^\top ] = \sum_{k=1}^t \decay^k \gossip^{2k} \expect [ \rwn\at{t-k}{\rwn\at{t-k}}^\top] = \sum_{k=1}^t \decay^k \gossip^{2k}.
    \end{align*}
    Using commutativity of $\gossip$ and its Eigen decomposition (Assumptions~\ref{assumption:symmetric}, \ref{assumption:doubly_stochastic}), we can decompose it as
    \begin{align*}
        \rwcov\at{t} = \sum_{i=1}^\nwrk \weigvec_i \weigvec_i^\top \underbrace{\left(\sum_{k=1}^t \decay^{k-1} \weigval_i^{2k} \right)}_{\rwcoveig\atidx{t}{i}}
    \end{align*}
    Because all terms of parenthesized expression are non-negative, and its limit equals $\frac{\weigval_i^2}{1 - \decay \weigval_i^2}$, this proves the Lemma.
\end{proof}

\begin{lemma}
    \label{lemma:rwalk-var}
    When the topology is regular (Assumption~\ref{assumption:regular}) in addition to the assumptions of Lemma~\ref{lemma:rwalk-cov}, workers in the random walk process have equal variance:
    \begin{align*}
        \Var[\rw\atidx{t}{i}] = \frac{1}{n} \Tr[\rwcov\at{t}] = \frac{1}{\nwrk}\sum_{i=1}^\nwrk \rwcoveig\atidx{t}{i}.
    \end{align*}
\end{lemma}
\begin{proof}
    The variances of $\rw_i$ are the diagonal entries of the covariance matrix.
    By regularity, and because workers are initialized equally, all workers should have the same variance.
    $\Var[\rw\atidx{t}{i}]$ is therefore equal to the average diagonal entry of $\rwcov\atidx{t}{i}$.
    The second equality is a standard property of the trace.
\end{proof}

\begin{lemma}
    \label{lemma:var-growing}
    Under the assumptions of Lemma~\ref{lemma:rwalk-var}, the variance $\Var[\rw\atidx{t}{i}]$ increases over time:
    \begin{align*}
    \Var[\rw\atidx{t}{i}] \leq \Var[\rw\atidx{t+1}{i}] \leq \lim_{t'\to\infty} \Var[\rw\atidx{t'}{i}] \qquad \forall t.
    \end{align*}
\end{lemma}
\begin{proof}
    If we write $\Var[\rw\atidx{t}{i}]$ as $\frac{1}{\nwrk}\sum_{i=1}^\nwrk \rwcoveig\atidx{t}{i}$ using Lemma~\ref{lemma:rwalk-var}, the statement of this Lemma follows from the realization in Lemma~\ref{lemma:rwalk-cov} that $\rwcoveig\atidx{t}{i}$ increases over time to the limit $\frac{\weigval_i^2}{1 - \decay \weigval_i^2}$ for all $i$.
\end{proof}

Note that while the results above are for static gossip matrices, random walks and these variance quantities can be analogously defined time-varying topologies.
Those just lack a simple exact form.
The stronger the averaging of the gossip process, the lower the variance.
We capture this in the following quantity:

\vspace{2mm}
\begin{definition}[Effective number of neighbors]
\label{def:eff-nn}
    \begin{align*}
        n_\gossip(\decay) = \frac{\frac{1}{1-\decay}}{ \lim_{t\to\infty} \frac{1}{n}\sum_{i=1}^\nwrk \Var[{\rw\atidx{t}{i}}]},
    \end{align*}
    where $\rw$ are the iterates from a random walk with gossip averaging, with decay parameter $\decay$.
    The numerator is the variance of the random walk process without any gossip averaging ($\mW=\mI$).
\end{definition}

\subsection{Converging random walk}

The covariance of the random walk process $\rwcov$ and the error matrix of \dsgd iterates $\emat$ share clear similarities.
The quantities are both iteratively updated by an affine transformation.
The main difference between them, however, is that $\rwcov$ converges to a non-zero constant while $\emat$ converges linearly to zero (or it diverges.)

In the next section, we draw a clear connection between the two processes, but first, we define a modified version of the random walk process that further highlights their similarity.

\vspace{2mm}
\begin{definition}[Scaled random walk]
    \label{def:scaled-random-walk}
    Let $0 < \rate < 1$ be a scalar.
    We define a scaled version of the random walk iterates, such that
    \begin{align*}
        \rws\at{t} &= (1-\rate)^{t/2}\rw\at{t}, \\
        \rwcovs\at{t} &= (1-\rate)^t \rwcov\at{t}, \text{ and} \\
        \Var[\rws\atidx{t}{i}] &= (1-\rate)^t \Var[\rw\atidx{t}{i}]
    \end{align*}
\end{definition}

Because the sequence $\rw\at{t}$ converges to a non-zero stationary point, the scaled sequence $\rws\at{t}$ converges to zero with a linear rate $\rate$.

\vspace{2mm}
\begin{lemma}
    \label{lemma:var-bound}
    Under the assumptions of Lemma~\ref{lemma:rwalk-var},
    the variance $\Var[\rws\atidx{t}{i}]$ is bounded as
    \begin{align*}
        \Var[\rws\atidx{t}{i}] \leq \frac{(1-\rate)^t}{(1-\decay)n_\gossip(\decay)},
    \end{align*}
    with equality as $t \to \infty$.
\end{lemma}
\begin{proof}
    From Lemma~\ref{lemma:var-growing}, we know that $(1-\rate)^t \Var[\rw\atidx{t}{i}] \leq (1-\rate)^t \lim_{t'\to\infty} \Var[\rw\atidx{t'}{i}]$, with equality as $t \to \infty$. Because the variance $\Var[\rw\atidx{t}{i}]$ is equal across workers $i$ (Lemma~\ref{lemma:rwalk-var}), the Lemma follows from rearranging Definition~\ref{def:eff-nn}.
\end{proof}

\vspace{2mm}
\begin{lemma}
    \label{lemma:bkwefoijwef}
    The covariance vector $\rwcovvs$ (the flattened version of $\rwcovs$) of this scaled random walk process follows the recursion $\rwcovvs\at{t+1} = \trans^\text{gossip} u_t(\rwcovvs\at{t})$, where
    \begin{align*}
        u_t(\rwcovvs\at{t})_{\nwrk i + j} = \begin{cases}
            \decay (1 - \rate) \, \rwcovvs_{\nwrk i + j}\at{t} + (1-\rate)^{t+1} & i = j \text{ (same worker)}, \\
            \decay (1 - \rate) \, \rwcovvs_{\nwrk i + j}\at{t} & i \neq j \text{ (different workers)}.
        \end{cases}
    \end{align*}
\end{lemma}
\begin{proof}
    The entries $u_t(\rwcovvs\at{t})_{\nwrk i + j}$ are inner products:
    \begin{align*}
        u_t(\rwcovvs\at{t})_{\nwrk i + j} &= (1-\rate)^{t+1}\lin{ \sqrt{\decay} \rws\atidx{t}{i} + \rwn\atidx{t}{i}, \sqrt{\decay} \rws_j\at{t} + \rwn_j\at{t}} \\
         &= \decay(1-\rate) \rwcovvs\at{t}_{\nwrk i + j} + (1-\rate)^{t+1} \expect \lin{\rwn\atidx{t}{i}, \rwn_j\at{t}}.
    \end{align*}
    The inner product between noise contributions $\rwn\atidx{t}{i}$ and $\rwn_j\at{t}$ are 1 if $i = j$ and 0 otherwise.
\end{proof}

\vspace{2mm}
\begin{lemma}
    \label{lemma:oiwjefoij}
    The covariance $\rwcovvs$ follows the recursion  $\rwcovvs\at{t+1} \geq \trans^\text{gossip} \trans^\text{r.w.}\rwcovvs\at{t}$ (element-wise), where
    \begin{align}
        \label{eq:recursion}
        [\trans^\text{r.w.}\rwcovvs\at{t}]_{\nwrk i + j} = \begin{cases}
            \decay (1 - \rate) \, \rwcovvs_{\nwrk i + j}\at{t} + (1-\rate) (1-\decay) n_\gossip(\decay) \rwcovvs_{\nwrk i + j}\at{t} & i = j, \\
            \decay (1 - \rate) \, \rwcovvs_{\nwrk i + j}\at{t} & i \neq j.
        \end{cases}
    \end{align}
    In the limit of $t \to \infty$, this is true with equality.
    \end{lemma}
    \begin{proof}
        From Lemma~\ref{lemma:var-bound}, we have that $\rwcovvs_{\nwrk i + i}\at{t} = \Var[\rws\atidx{t}{i}] \leq \frac{(1-\rate)^t}{(1-\decay)n_\gossip(\decay)}$, with equality as $t \to \infty$.
        The entries of $\trans^\text{r.w.}\rwcovvs\at{t}$ are therefore all smaller than or equal to the entries of $u_t(\rwcovvs\at{t})$ from Lemma~\ref{lemma:bkwefoijwef}, which proves the Lemma.
    \end{proof}

\subsection{The rate for \dsgd}

\begin{theorem}[\dsgd on random quadratics]
    Under assumptions \ref{assumption:constant}, \ref{assumption:symmetric}, \ref{assumption:doubly_stochastic}, and \ref{assumption:regular},
    if the pair of the learning rate $\lr$ and $\rate$ satisfy
    \begin{align}
        \label{eqn:condition}
        \rate = 1 - (1-\lr)^2 - \frac{(\noise - 1) \lr^2}{n_\gossip\left(\frac{(1-\lr)^2}{1 - \rate}\right)},
    \end{align}
    the error of \dsgd with learning rate $\lr$ on the random quadratic objective with noise parameter $\noise$ converges with rate $\rate$:
    \begin{align*}
        \sum_{i=1}^\nwrk \expect \norm{\param\atidx{t}{i}}^2 \leq (1-\rate)^t \sum_{i=1}^\nwrk \expect \norm{\param\idx{i}\at{0}}^2.
    \end{align*}
    This rate becomes exact as $t \to \infty$.
\end{theorem}
\begin{proof}
    If the condition \eqref{eqn:condition} is satisfied, the expected error iterates $\emat$ (Equation \ref{eqn:cov-sgd}) of the \dsgd algorithm follow the transition matrix \eqref{eq:recursion} of Lemma~\ref{lemma:oiwjefoij} with $\decay = \frac{(1-\lr)^2}{1-\rate}$.
    The choice of $\decay$ ensures that $\decay(1-\rate) = (1-\lr)^2$, and the condition \eqref{eqn:condition} that $(\noise - 1)\eta^2 = (1-\rate)(1-\decay)n_\gossip(\decay)$.

    From Lemma~\ref{lemma:oiwjefoij}, we know that a sequence that has this transition matrix $\trans^\text{gossip} \trans^\text{r.w.}\rwcovvs\at{t}$ \eqref{eq:recursion} converges at least as fast as the iterates of the corresponding scaled random walk process $\rwcovs$, with equality in the limit as $t \to \infty$.

    Since $\rwcovs$ converges to zero with a rate $\rate$, this now implies the same rate for the error matrix $\emat$. The sum $\sum_{i=1}^\nwrk \expect \norm{\param\atidx{t}{i}}^2$ in the statement of this theorem is the trace of the matrix $\emat\at{t}$, and therefore it converges to zero with the same rate.
    This completes the proof.
\end{proof}

%% file: 095_appendix_20_theory.tex
\clearpage
\section{(Strongly)-Convex case, missing proofs and additional results}
\label{apx:theory-convex}

\subsection{Preliminaries on Bregman divergences}
Throughout this section, we will use Bregman divergences, which are defined for a differentiable function $h$ and two points $x,y \in \R^d$ as:
\begin{equation}
    D_h(x,y) = h(x) - h(y) - \nabla h(y)^\top(x - y).
\end{equation}
We assume throughout this paper that the functions we consider are twice continuously differentiable and strictly convex on ${\rm dom}\;h$, and that $\nabla h
(x) = \min_y h(y) - x^\top y$ is uniquely defined (although milder assumptions could be used). Among the many properties of these divergences, an important one is that if $h$ is $\smoothf$ smooth and $\mu$ strongly-convex, then
\begin{equation}
    \frac{\mu}{2}\norm{x - y}^2 \leq D_h(x, y) \leq \frac{\smoothf}{2}\norm{x - y}^2
\end{equation}
Another important property is called duality, which states that:
\begin{equation}
    D_h(x, y) = D_{h^*}(\nabla h(y), \nabla h(x)),
\end{equation}
where $h^*$ is the convex conjugate of $h$.

\subsection{Main result}
This section is devoted to proving Theorem~\ref{thm:convex_precise}, from which Theorem~\ref{thm:convex_rates} can be deduced directly by taking $\wcons = M_0$ and $p = 1/2$. We recall Assumption~\ref{assumption:stochastic}, which is at the heart of Theorem~\ref{thm:convex_precise}.

\begin{nonumber_assumption}
	The stochastic gradients are such that: (\textsc{i}) $\xi\atidx{t}{i}$ and $\xi\atidx{\ell}{j}$ are independent for all $t, \ell$ and $i \neq j$.
	(\textsc{ii}) $\expect{[ f_{\xi\atidx{t}{i}}]} = f$ for all $t, i$
	(\textsc{iii}) $\expect{\norm{\nabla f_{\xi\atidx{t}{i}}(\param\at{\star})}^2} \leq \variance^2$ for all $t, i$, where $\param\at{\star}$ is a minimizer of $f$.
	(\textsc{iv}) $f_{\xi\atidx{t}{i}}$ is  convex and $\noise$-smooth for all $t,i$.
	(\textsc{v}) $f$ is $\mu$-strongly-convex for $\mu\ge0$ and $\smoothf$-smooth.
\end{nonumber_assumption}

Note that Assumption B (IV) is stated in this form for simplicity, but it can be relaxed by asking directly that $\esp{\| \nabla f_{\xi,i}(x^{(t)}) - \nabla f_{\xi,i}(x^\star)\|^2} \leq 2 \zeta D_f(x^\star, x^{(t)})$, which can also be implied by assuming that each $f_\xi$ is $\zeta_\xi$-smooth, with $\esp{\zeta_\xi D_{f_\xi}(x^\star, x^{(t)})} \leq \zeta D_f(x^\star, x^{(t)})$ (see Equation~\eqref{eq:relaxed_smoothness}). These weaker forms would be satisfied by the toy problem of Section~\ref{sec:toy-model}.

\begin{theorem}\label{thm:convex_precise}
    Denote $\param\at{t}$ the iterates obtained by \dsgd, $\LM = \mI - \mmat$, and $p$ the probability to perform a communication step ($\param_{t+1} = \gossip \param_t$). Parameter $\beta$ is such that $\mI - \gossip \succcurlyeq \beta \LM$. For some $\wcons > 0$, denote:
    \begin{equation}
        \cL_t = \norm{\param\at{t} - \param\at{\star}}_\mmat^2 + \wcons \norm{ \param\at{t} }^2_{\LM}.
    \end{equation}
    Then, if $\lr$ is such that:
    \begin{align}
        & \lr \leq \frac{M_0\beta}{\smoothf} \frac{p}{1 - p}, \label{eq:eta_comm_constraint}\\
        & \lr \leq \frac{1}{4\left(M_0 \noise + \smoothf\right)} \label{eq:eta_comp_constraint}
    \end{align}
    we have that:
    \begin{equation}
        \cL_t \leq [1 - (1-p)\lr \mu]^t \cL_0 + \frac{\lr \mvar}{\mu},
    \end{equation}
    with $\mvar = \variance_\mmat^2 + \wcons \variance_{\LM}^2$, where $\expect{\norm{\nabla f_{\xi\atidx{t}{i}}(\param\at{\star})}^2_\mmat} \leq \variance^2_\mmat$ (and similarly for $\LM$).

    In the convex case ($\mu = 0$), we have:
    \begin{equation}
        \esp{\frac{1}{T} \sum_{t=0}^{T-1} D_f(\mmat \param\at{t}, \param\at{\star})} \leq \frac{1}{1 - p}\frac{\cL_0}{\lr T} + \lr \mvar
    \end{equation}
\end{theorem}

Note that the factors $2$ in Equation~\eqref{eq:eta_comp_constraint} are simplifications to make the result more readable but could be improved.

\begin{proof}
We now proceed to the proof of the theorem. To show that the Lyapunov $\cL_t$ decreases over iterations, we will study how each quantity $\norm{\param\at{t} - \param\at{\star}}_\mmat^2$ and $\norm{ \param\at{t}}^2_{\LM}$ evolves through time. In particular, we will first consider the case of computation updates (so, local gradient updates), and then the case of gossip updates.

\paragraph{1 - Computation updates}
In this case, we assume that the update is of the form
\begin{equation}
    \param\at{t+1} = \param\at{t} - \lr \nabla f_{\xi_t}(\param\at{t}).
\end{equation}
This happens with probability $1-p$, and expectations are taken with respect to $\xi_t$. To avoid notation clutter, we use notations $\nabla f_\xi$ and $\nabla f_{\xi, i}$, which are such that $\nabla f_{\xi, i}(\param^{(t)}) = (\nabla f_\xi(\param^{(t)}))_i = \nabla f_{\xi_i^{(t)}}(\param_i^{(t)})$.

\paragraph{Distance to optimum}
We bound the distance to optimum as follows, using that $\mmat \param\at{\star} = \param\at{\star}$, and $\esp{\nabla f_\xi(\param\at{t})} = \nabla f(\param\at{t})$:
\begin{align*}
    \esp{\norm{\param\at{t+1} - \param\at{\star}}^2_\mmat} &= \norm{\param\at{t} - \param\at{\star}}^2_\mmat - 2 \lr \esp{(\param\at{t} - \param\at{\star})^\top \mmat \nabla f_\xi(\param\at{t})} + \lr^2 \norm{\nabla f_\xi(\param\at{t})}^2_\mmat\\
    &= \norm{\param\at{t} - \param\at{\star}}^2_\mmat - 2 \lr (\mmat \param\at{t} - \param\at{\star})^\top \nabla f(\param\at{t}) + \lr^2 \esp{\norm{\nabla f_\xi(\param\at{t})}^2_\mmat}.
\end{align*}
Then, we expand the middle term in the following way:
\begin{align}
    - \nabla f(\param\at{t})^\top(\mmat \param\at{t} - \param\at{\star})  &= - \nabla f(\param\at{t})^\top(\param\at{t} - \param\at{\star}) - \nabla f(\param\at{t})^\top(\mmat \param\at{t} - \param\at{t}) \nonumber\\
    &= - D_f(\param\at{t}, \param\at{\star}) - D_f(\param\at{\star}, \param\at{t}) + D_f(\mmat \param\at{t}, \param\at{t}) - f(\mmat \param\at{t}) + f(\param\at{t})\nonumber \\
    &= - D_f(\mmat \param\at{t}, \param\at{\star}) - D_f(\param\at{\star}, \param\at{t}) + D_f(\mmat \param\at{t}, \param\at{t}) \nonumber\\
    &\leq - \frac{\mu}{2}\norm{\param\at{t} - \param\at{\star}}^2_{\mmat^2} - D_f(\param\at{\star}, \param\at{t}) + \frac{\smoothf}{2}\norm{\mmat \param\at{t} - \param\at{t}}^2 \label{eq:proof_sc},
\end{align}
where in the last time we used the $\mu$-strong convexity and $\smoothf$-smoothness of $f$. For the noise term, we use that fact that $(\nabla f_\xi (\param\at{t}))_i$ and $(\nabla f_\xi (\param\at{t}))_j$ are independent for $i \neq j$, so that
\begin{align*}
    \frac{1}{2}\esp{\norm{\nabla f_\xi(\param\at{t})}^2_\mmat} &= \esp{\norm{\nabla f_\xi(\param\at{t}) - \nabla f_\xi(\param\at{\star})}^2_\mmat} + \esp{\norm{\nabla f_\xi(\param\at{\star})}^2_\mmat}\\
    &= \sum_{i=1}^n \mweight{ii} \esp{\norm{\nabla f_{\xi,i}(\param\at{t}) - \nabla f_{\xi,i}(\param\at{\star})}^2} + \esp{\norm{\nabla f_\xi(\param\at{\star})}^2_\mmat}\\
    & + \sum_{i=1}^n \sum_{j \neq i} \mweight{ij} \esp{[\nabla f_{\xi,i}(\param\at{t}) - \nabla f_{\xi,i}(\param\at{\star})]^\top [\nabla f_{\xi,j}(\param\at{t}) - \nabla f_{\xi,j}(\param\at{\star})]}\\
    &= \sum_{i=1}^n \mweight{ii} \esp{\norm{\nabla f_{\xi,i}(\param\at{t}) - \nabla f_{\xi,i}(\param\at{\star})}^2} + \norm{\nabla f (\param\at{t})}^2_{\mmat} + \esp{\norm{\nabla f_\xi(\param\at{\star})}^2_\mmat}.
\end{align*}
We now use for all $i \in \{1, \dots, n\}$ the $\noise$-smoothness of $f_{\xi,i}$, which implies the $\noise^{-1}$-strong convexity of $f_{\xi,i}^*$~\citep{kakade2009duality}, so that:
\begin{align}
    \esp{\norm{\nabla f_{\xi,i}(\param\at{t}) - \nabla f_{\xi,i}(\param\at{\star})}^2} &= 2\esp{D_{\frac{1}{2}\norm{ \cdot }^2}(\nabla f_{\xi,i}(\param\at{t}), \nabla f_{\xi,i}(\param\at{\star}))} \nonumber\\
    &\leq \frac{2}{\noise^{-1}} \esp{D_{f^*_{\xi,i}}(\nabla f_{\xi,i}(\param\at{t}),\nabla f_{\xi,i}(\param\at{\star}))}\nonumber\\
    &\leq 2 \noise \esp{D_{f_{\xi,i}}(\param\at{\star}, \param\at{t})} \label{eq:relaxed_smoothness}\\
    &= 2 \noise D_{f}(\param\at{\star}, \param\at{t})\nonumber
\end{align}
For the expected gradient term, we can use that:
\begin{align*}
    \norm{\nabla f (\param\at{t})}^2_{\mmat} \leq \norm{\nabla f (\param\at{t}) - \nabla f(\param\at{\star})}^2 \leq 2 \smoothf D_f(\param\at{\star}, \param\at{t}),
\end{align*}
so that in the end,
\begin{equation}
    \esp{\norm{\nabla f_\xi(\param\at{t})}^2_\mmat} \leq 4(\noise M_0 + \smoothf) D_f(\param\at{\star}, \param\at{t}) + 2\variance_\mmat^2,
\end{equation}
where $M_0 = \max_i \mweight{ii}$, and $\esp{\norm{\nabla f_\xi(\param\at{\star})}^2_\mmat} \leq \variance_\mmat^2$, the locally averaged variance at optimum. Plugging this into the main equation, we obtain that:
\begin{align*}
    \esp{\norm{\param\at{t+1} - \param\at{\star}}^2_\mmat} &\leq \norm{\param\at{t} - \param\at{\star}}^2_\mmat - \lr \mu\norm{\param\at{t} - \param\at{\star}}^2_{\mmat^2} + 2\lr^2 \variance_\mmat^2\\
    &- 2\lr \left(1 - 2 \lr \left[\noise M_0 + \smoothf\right]\right)D_f(\param\at{\star}, \param\at{t}) + \frac{\smoothf}{2}\norm{\mmat \param\at{t} - \param\at{t}}^2
\end{align*}
The last step is to write that $\mmat^2 = \mmat - \mmat \LM$, so that:
\begin{align*}
    \esp{\norm{\param\at{t+1} - \param\at{\star}}^2_\mmat} &\leq (1 - \lr \mu)\norm{\param\at{t} - \param\at{\star}}^2_\mmat + \lr \mu\norm{\param\at{t} - \param\at{\star}}^2_{\mmat \LM} + 2 \lr^2 \variance_\mmat^2\\
    &- 2\lr \left(1 - 2\lr \left[\noise M_0 + \smoothf\right]\right)D_f(\param\at{\star}, \param\at{t}) + \lr \smoothf\norm{\mmat \param\at{t} - \param\at{t}}^2
\end{align*}
At this point, we can use that $\mmat \LM \leq \mI / 4$,
\begin{equation}
    \mu\norm{\param\at{t} - \param\at{\star}}^2_{\mmat \LM} \leq \frac{\mu}{4}\norm{\param\at{t} - \param\at{\star}}^2 \leq \frac{1}{2}D_f(\param\at{\star}, \param\at{t}),
\end{equation}
so that
\begin{equation}
    \begin{split} \label{eq:main_comp_suboptimality}
        \esp{\norm{\param\at{t+1} - \param\at{\star}}^2_\mmat} &\leq (1 - \lr \mu)\norm{\param\at{t} - \param\at{\star}}^2_\mmat +  \lr \smoothf\norm{\mmat \param\at{t} - \param\at{t}}^2\\
    &- 2\lr \left(3/4 - 2\lr \left[\noise M_0 + \smoothf\right]\right)D_f(\param\at{\star}, \param\at{t}) + 2 \lr^2 \variance_\mmat^2
    \end{split}
\end{equation}

\paragraph{Distance to consensus}
We now bound the distance to consensus in the case of a communication update. More specifically, we write that:
\begin{align*}
    \esp{\norm{\param\at{t+1}}^2_{\LM}} &= \norm{\param\at{t}}^2_{\LM} - 2 \lr \nabla f(\param\at{t})^\top \LM \param\at{t} + \lr^2 \esp{\norm{\nabla f_{\xi_t}(\param\at{t})}^2_{\LM}}
\end{align*}
Then, we develop the middle term as:
\begin{align*}
	- \nabla f(\param\at{t})^\top \LM \param\at{t} &= - \nabla f(\param\at{t})^\top (\mI - \mmat) \param\at{t}\\
	&=  \nabla f(\param\at{t})^\top ( \mmat \param\at{t} - \param\at{t})\\
	&= - D_f(\mmat \param\at{t}, \param\at{t}) + f(\mmat \param\at{t}) - f(\param\at{t})\\
	&\leq - \frac{\mu}{2} \norm{\mmat \param\at{t} - \param\at{t}}^2 + f(\mmat \param\at{t}) - f(\param\at{t})
\end{align*}
By convexity of $f$ (since the expected function is the same for all workers), we have that
\begin{equation}
	f(\mmat \param\at{t}) \leq f(\param\at{t}).
\end{equation}
We finally decompose $\LM^2 = \LM (\mI - \mmat)$, so that:
\begin{align*}
- 2 \lr \nabla f(\param\at{t})^\top \LM \param\at{t} &\leq - \lr \mu\norm{\param\at{t} - \param\at{\star}}^2_{\LM} + \lr \mu \norm{\param\at{t}}^2_{\mmat \LM}
\end{align*}
For the noise term, we obtain exactly the same derivations as in the previous setting, but this time with matrix $\LM = \mI - \mmat$ instead. Using the same bounding, and $M_{\min} = \min_i (\mmat)_{ii}$, we thus obtain:
\begin{equation}
    \esp{\norm{\nabla f_\xi(\param\at{t})}^2_{\LM}} \leq 4(\noise(1 - M_{\min}) + \smoothf) D_f(\param\at{\star}, \param\at{t}) + 2 \variance^2_{\LM}.
\end{equation}
In particular, we have that:
\begin{align*}
    \esp{\norm{\param\at{t+1} - \param\at{\star}}^2_{\LM}} &\leq (1 - \lr \mu)\norm{\param\at{t} - \param\at{\star}}^2_{\LM} + \lr \mu \norm{\param\at{t}}^2_{\mmat \LM} + 2\lr^2 \variance^2_{\LM}\\
    &+ 4\lr^2(\noise(1 - M_{\min}) + \smoothf) D_f(\param\at{\star}, \param\at{t}).
\end{align*}
Similarly to before, we use that
\begin{equation}
    \mu \norm{\param\at{t}}^2_{\mmat\LM} = \mu \norm{\param\at{t} - \param\at{\star}}^2_{\mmat\LM} \leq \frac{\mu}{4} \norm{\param\at{t} - \param\at{\star}}^2 \leq \frac{1}{2}D_f(\param\at{\star}, \param\at{t}),
\end{equation}
so that for computation updates, the distance to consensus evolves as:
\begin{equation} \label{eq:main_comp_consensus_dist}
    \esp{\norm{\param\at{t+1} - \param\at{\star}}^2_{\LM}} \leq (1 - \lr \mu)\norm{\param\at{t} - \param\at{\star}}^2_{\LM} + 2\lr\left[ \frac{1}{4} + 2 \lr (\noise(1 - M_{\min}) + \smoothf)\right] D_f(\param\at{\star}, \param\at{t}) + 2\lr^2 \variance_{\LM}^2
\end{equation}
Combining Equation~\eqref{eq:main_comp_consensus_dist} with Equation~\eqref{eq:main_comp_suboptimality} leads to:
\begin{equation}\label{eq:master_comp}
    \begin{split}
        \cL\at{t+1} &\leq (1 - \lr \mu)\cL_t +  \lr \smoothf\norm{\mmat \param\at{t} - \param\at{t}}^2 + 2\lr^2 \mvar\\
        &- \lr \left(1 - 4\lr \left[\noise (M_0 + \wcons(1 - M_{\min})) + (1 + \wcons)\smoothf\right]\right)D_f(\param\at{\star}, \param\at{t}),
    \end{split}
\end{equation}
with $\mvar = \variance^2_\mmat + \wcons \variance_{\LM}^2$.

\paragraph{2 - Communication updates}
We write:
\begin{align*}
    \norm{\param\at{t+1} - \param\at{\star}}^2_{\LM} &= \norm{\param\at{t} - \param\at{\star}}^2_{\gossip \LM \gossip} \\
    &\leq \norm{\param\at{t} - \param\at{\star}}^2_{\gossip \LM} \\
    &= \norm{\param\at{t} - \param\at{\star}}^2_{\LM} - \norm{\param\at{t} - \param\at{\star}}^2_{\LW \LM}
\end{align*}
For distance to optimum part in communication update, we obtain:
\begin{equation}
    \norm{\param\at{t+1} - \param\at{\star}}^2_{\mmat} = \norm{\param\at{t} - \param\at{\star}}^2_{\gossip \mmat \gossip} \leq \norm{\param\at{t} - \param\at{\star}}^2_{\mmat}
\end{equation}
We now introduce $\beta$, the strong convexity of $\LW = \mI - \gossip$ relative to $\LM$:
\begin{equation}
    \LW \geq \beta \LM.
\end{equation}
Therefore, we obtain that for communication updates,
\begin{equation} \label{eq:master_comm}
    \cL\at{t+1} \leq \cL_t  - \wcons \beta \norm{\param\at{t} - \param\at{\star}}^2_{\LM^2}.
\end{equation}

\paragraph{Putting terms back together}
We now put everything together, assuming that communication steps happen with probability $p$ (and so computations steps with probability $1 - p$). Thus, we mix Equations~\eqref{eq:master_comp} and~\eqref{eq:master_comm} to obtain:
\begin{align*}
    \esp{\cL\at{t+1}} &\leq  (1 - (1 - p)\lr\mu) \cL_t + 2(1-p)\lr^2 \mvar\\
    &  + \left[(1 - p)\lr \smoothf - \wcons p \beta \right] \norm{\param\at{t}}^2_{\LM^2} \\
    &- \lr (1 - p)\left(1 - 4\lr \left[\noise (M_0 + \wcons(1 - M_{\min})) + (1 + \wcons)\smoothf\right]\right)D_f(\param\at{\star}, \param\at{t}).
\end{align*}

In particular, we obtain the linear decrease of the Lyapunov $\cL_t$ under the following conditions:
\begin{align*}
    &\lr \leq \frac{\wcons\beta}{\smoothf} \frac{p}{1 - p}\\
    &\lr \leq \frac{1}{4\left(\noise \left[M_0 + \wcons (1 -M_{\min})\right] + (1 + \wcons) \smoothf\right)}
\end{align*}

Under these conditions, we have that
\begin{equation}
    \esp{\cL\at{t+1}} \leq  (1 - (1 - p)\lr\mu) \cL_t + (1 - p)\lr^2 \mvar,
\end{equation}
and we can simply chain this relation to finish the proof of the theorem.
\end{proof}

\paragraph{Convex case}
In the convex case ($\mu = 0$), the proof is very similar, except that we keep the $D_f(\mmat \param\at{t}, \param\at{\star})$ term from Equation~\eqref{eq:proof_sc}. In particular, under the same step-size conditions as the strongly convex case, this leads to:
\begin{equation}
    \esp{\cL\at{t+1}} \leq \cL_t + 2(1-p)\lr^2 \mvar - \lr (1 - p)D_f(\mmat \param\at{t}, \param\at{\star}).
\end{equation}
This leads to:
\begin{equation}
    \esp{\frac{1}{T} \sum_{t=0}^{T-1} D_f(M \param\at{t}, \param\at{\star})} \leq \frac{1}{1 - p}\frac{\cL_0}{\lr T} + 2\lr \mvar,
\end{equation}
which finishes the proof of the theorem.

\paragraph{Evaluating $\beta$}
There are two important graph quantities: $M_0$ and $\beta$. If we choose $M$ as in Equation~\eqref{eq:M_definition}, then its eigenvalues are equal to $\frac{(1 - \decay)\weigval_i^2}{1 - \decay\weigval_i^2}$, where $\weigval_i$ is the $i$-th eigenvalue of $\gossip$. Therefore,
\begin{equation}
    \weigval_i^{\LM} = \frac{1 - \weigval_i^2}{1 - \decay \weigval_i^2}.
\end{equation}
In particular, we have that for all $i$,
\begin{equation}
    1 - \weigval_i \geq \beta \frac{1 - \weigval_i^2}{1 - \decay \weigval_i^2},
\end{equation}
so that we can take
\begin{equation}
    \beta = \frac{1 - \decay \lmax^2}{1 + \lmax} \geq \frac{1 - \decay \lmax}{2},
\end{equation}
where we use $\lmax \leq 1$ to simplify the results.
In particular, $\beta$ does not depend on the spectral gap of $\gossip$ (which is equal to $1 - \lmax$) as long as $\decay$ is not too large. Yet, an interesting phenomenon happens: \textit{a larger graph also implies more effective neighbors for a given $\decay$}. %

\paragraph{Choice of $\wcons$}
A reasonable value for $\wcons$ is to simply take it as $\wcons = M_0$. Indeed,
\begin{itemize}
    \item The second condition almost does not benefit from $\wcons \leq M_0$ (factor $2$ at most).
    \item If the first condition dominates, such that taking $\wcons \geq M_0$ would loosen it, then instead one can reduce $\decay$. This will lead to a higher value for both $M_0$ (and so for $\wcons$) and $\beta$. Note that, again, increasing $M_0$ does not make the second condition stronger than what it would have been with just increasing $\wcons$ by more than a factor $2$.
\end{itemize}
With this choice, we thus obtain that:
\begin{equation}
    \lr \leq \min \left( \frac{M_0\beta}{\smoothf} \frac{p}{1 - p}, \frac{1}{4\left( M_0 \noise (2 - M_{\min}) + (1 + M_0)\smoothf\right)} \right),
\end{equation}
and Theorem~\ref{thm:convex_precise} is obtained by taking $M_0 \leq 1$ and $M_{\min} \geq 0$.

\subsection{Obtaining Corollary~\ref{corr:rates}}
In this section, we discuss the derivations leading to Corollary~\ref{corr:rates}.
To do so, we start by making the simplifying assumption that
\begin{equation}
    \frac{\noise}{n} \geq \smoothf.
\end{equation}
Using this, and writing $\neff = 1 / M_0$, the condition from Equation~\eqref{eq:eta_comp_constraint} simplifies to:
\begin{equation}
    \lr \leq \frac{\smoothf \neff}{16 \noise}.
\end{equation}
We always want this condition to be tight, and not Equation~\eqref{eq:eta_comm_constraint} the communication one, which is only there to allow us to use larger values of $\neff$. In particular, we want that:
\begin{equation} \label{eq:corr_simple}
    \frac{\smoothf \neff}{16 \noise} \leq \frac{\beta}{\neff \smoothf}.
\end{equation}
When we increase $\decay$, $\neff$ increases and $\beta$ decreases. We thus want to take the highest $\decay$ such that \eqref{eq:corr_simple} is verified (potentially with an equality if $\neff < n$).

\subsection{Deterministic algorithm}
\label{apx:deterministic-algorithm}
So far, we have analyzed the randomized variant of \dsgd, in which at each step, there is a coin flip to decide whether to perform a communication or computation step. We now show how to extend the analysis to the case in which:
\begin{equation}\label{eq:deterministic_d_sgd_new_form}
    \param\at{t+1} = \gossip\param\at{t} - \lr \nabla f_\xi(\gossip \param\at{t})
\end{equation}
Note that \dsgd is often presented as $\param_{t+1} = \gossip(\param_t - \eta \nabla f_\xi(\param_t))$, but it turns out that the analysis is easier when considering it in the form of Equation~\eqref{eq:deterministic_d_sgd_new_form}.
Yet, it comes down to the same algorithm (alternating communication and computation steps), and the difference simply is whether the error is evaluated after a communication step or a local gradient step.
The results in the previous section did not depend on the value of $\param_t$, so we can perform the same derivations with $\gossip\param_t$ instead of $\param_t$, so that Equation~\eqref{eq:master_comp} now writes:
\begin{equation}\label{eq:master_comp_W}
    \begin{split}
        \cL(\param\at{t+1}) &= (1 - \lr \mu)\cL(\gossip\param\at{t}) +  \lr \smoothf\norm{\mmat \gossip \param\at{t} - \gossip\param\at{t}}^2 + 2\lr^2 \mvar\\
        &- \lr \left(1 - 4\lr \left[\noise (M_0 + \wcons(1 - M_{\min})) + (1 + \wcons)\smoothf\right]\right)D_f(\param\at{\star}, \gossip\param\at{t}),
    \end{split}
\end{equation}
where $\cL(\param) = \norm{\param - \param\at{\star}}_\mmat^2 + \wcons \norm{ \param }^2_{\LM}$, so that $\cL\at{t} = \cL(\param\at{t})$. In particular, choosing $\lr$ such that the second line is always negative (as before) leads to:
\begin{equation}\label{eq:master_comp_W_simplified}
    \cL(\param\at{t+1}) = (1 - \lr \mu)\cL(\gossip\param\at{t}) +  \lr \smoothf\norm{\param\at{t}}^2_{\gossip \LM^2 \gossip} + 2\lr^2 \mvar.
\end{equation}
Similarly, using Equation~\eqref{eq:master_comm}, we obtain that
\begin{equation} \label{eq:master_comm_W}
    \cL(\gossip \param\at{t}) \leq \cL(\param\at{t})  - \wcons \beta \norm{\param\at{t} - \param\at{\star}}^2_{\LM^2}.
\end{equation}
Combining Equations~\eqref{eq:master_comp_W_simplified} and~\eqref{eq:master_comm_W} and using that $\gossip \LM^2 \gossip \preccurlyeq \LM^2$, we obtain:
\begin{equation}
    \cL\at{t+1} \leq (1 - \lr \mu)\cL\at{t} +  (\lr \smoothf - (1 - \lr \mu)\omega \beta)\norm{\param\at{t}}^2_{ \LM^2 } + 2\lr^2 \mvar.
\end{equation}
Thus, we obtain similar guarantees (up to a factor $1 - \eta \mu$ which is small) for the deterministic and randomized algorithms. Note that in this case, constant $\beta$ can be replaced by a slightly better constant $\tilde{\beta}$ which would be such that:
\begin{equation}
    \LM \LW \geq \tilde{\beta} \ \gossip \LM^2 \gossip.
\end{equation}

%% file: 095_appendix_70_cifar_details.tex
\clearpage
\section{Cifar-10 experimental setup}
\label{apx:cifar-details}

\autoref{tab:cifar-experimental-settings} describes the details of our experiments with \dsgd with VGG-11 on Cifar-10.

\begin{table}[h]
	\caption{
		Default experimental settings for Cifar-10/VGG-11
		\label{tab:cifar-experimental-settings}%
	}
	\scriptsize%
	\begin{tabularx}{\linewidth}{lX}
		\toprule
		Dataset              & Cifar-10~\citep{cifar10}                                                                                           \\
		Data augmentation    & Random horizontal flip and random $32\times 32$ cropping                                                           \\
		Data normalization    & Subtract mean $(0.4914, 0.4822, 0.4465)$ and divide standard deviation $(0.2023, 0.1994, 0.2010)$                                                           \\
		Architecture         & VGG-11~\citep{simonyan2015vgg}                                                                              \\
		Training objective   & Cross entropy                                                                                                      \\
		Evaluation objective & Top-1 accuracy                                                                                                     \\
		\midrule
		Number of workers    & 32 (unless otherwise specified)                                                                                                                 \\
		Topology             & Ring (unless otherwise specified) \\
		Gossip weights       & Metropolis-Hastings (1/3 for ring, $\gossipweight{ij} = 1 / (\max(n_i, n_j) + 1)$, worker $i$ has $n_i$ direct neighbors)                                                                                 \\
		Data distribution    & Identical: workers can sample from the whole dataset              \\
        Sampling    & With replacement (\iid), \emph{no} shuffled passes              \\
		\midrule
		Batch size           & 16 patches per worker                                                                                              \\
		Momentum             & 0.9 (heavy ball / PyTorch default)                                                                                                     \\
		Learning rate        & Exponential grid or tuned for lowest training loss after 25 epochs                                                                               \\
		LR decay             & Step-wise, $\times 0.1$ at epoch 75\% and 90\% of training                                                                                         \\
		LR warmup            & None                                                                \\
		\# Epochs            & 100 (full training) or only 25 (initial phase), based on total number of gradient accesses across workers                                                                                                                \\
		Weight decay         & $10^{-4}$                                                                                                          \\
		Normalization scheme & no normalization layers                                                                                             \\
        Exponential moving average  & $\param\atidx{t}{\text{ema}} = 0.95 \mspace{1mu} \param\atidx{t-1}{\text{ema}} + 0.05 \mspace{1mu} \param\at{t}$. This influences evaluation, not training \\
		\midrule
		Repetitions per training & Just 1 per learning rate, but experiments are very consistent across similar learning rates                                                                                                \\
		Reported metrics      & \emph{Loss after 2.5\kilo steps}:
            to reduce noise, we take two measures: (\textsc{i}) we use exponential moving average of the model parameters, and (\textsc{ii}) we fit a parametric model $\log(l)= a t + b$ to the 25 loss evaluations $(t, l)$ closest to $t=2500$.
            We then evaluate this function at $t=2500$.
            \\
		\bottomrule
	\end{tabularx}
\end{table}

%% file: 095_appendix_30_experiments.tex
\clearpage
\section{Additional experiments}
\label{apx:experiments}

\begin{figure}[h]
    \includegraphics{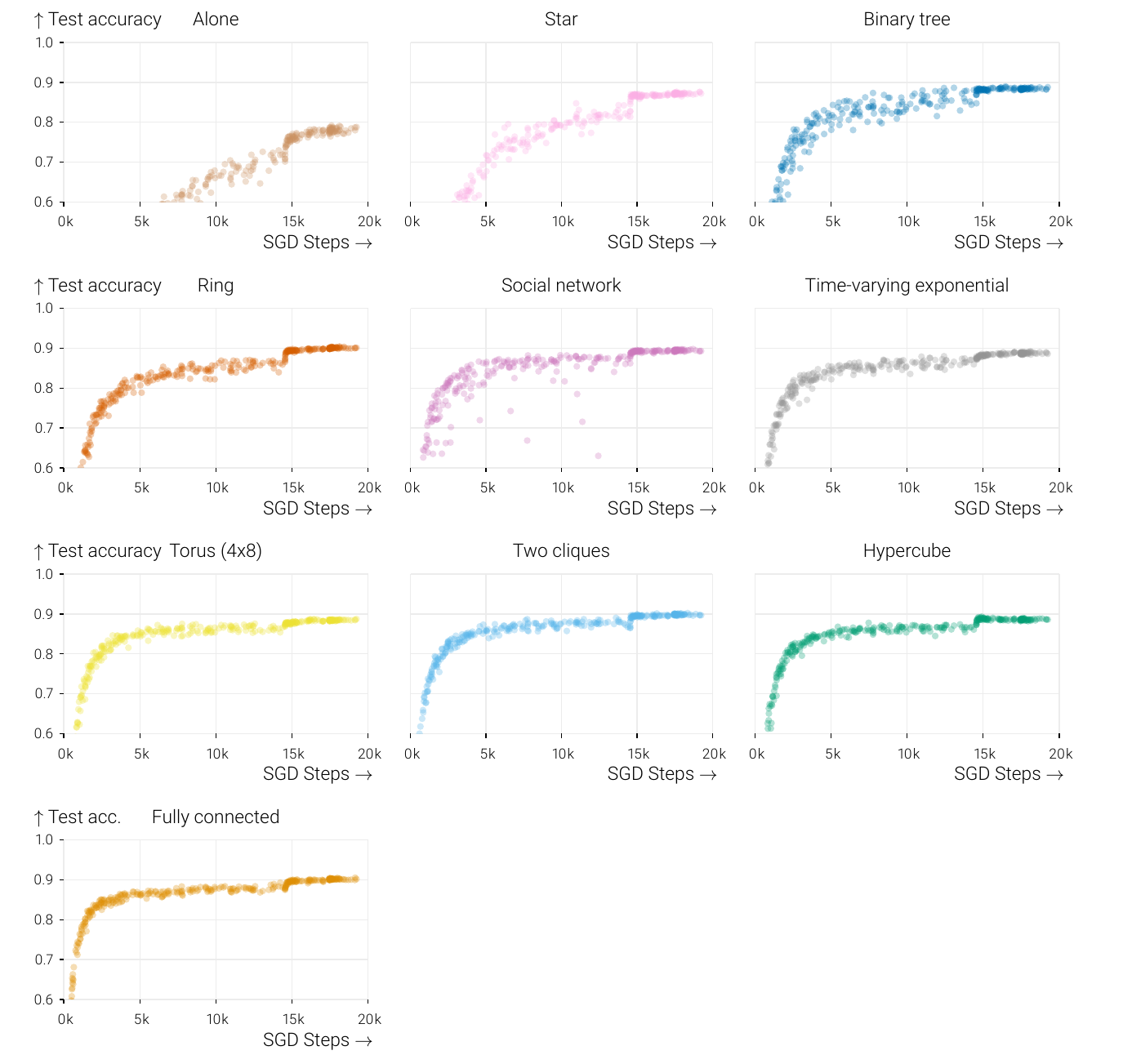}
    \caption{
        \label{apx:fig:cifar-full-result}
        Test accuracy over the course of training a VGG-11 network on Cifar-10.
        See \autoref{apx:cifar-details} for all details on the experimental setup.
        The plots are ordered by improving training loss after 2.5\kilo SGD steps.
        This ordering correlates well with the speed of improvements in test accuracy.
    }
\end{figure}

In the main paper, we have focussed on the training loss in the initial phase of training of Cifar-10.
We do find that our findings there do correlate with test accuracy after a complete training with 100 epochs.
\autoref{apx:fig:cifar-full-result} shows the test accuracy as training progresses, for plots ordered by improving training loss after 2.5\kilo steps.

\begin{figure}[h]
    \includegraphics{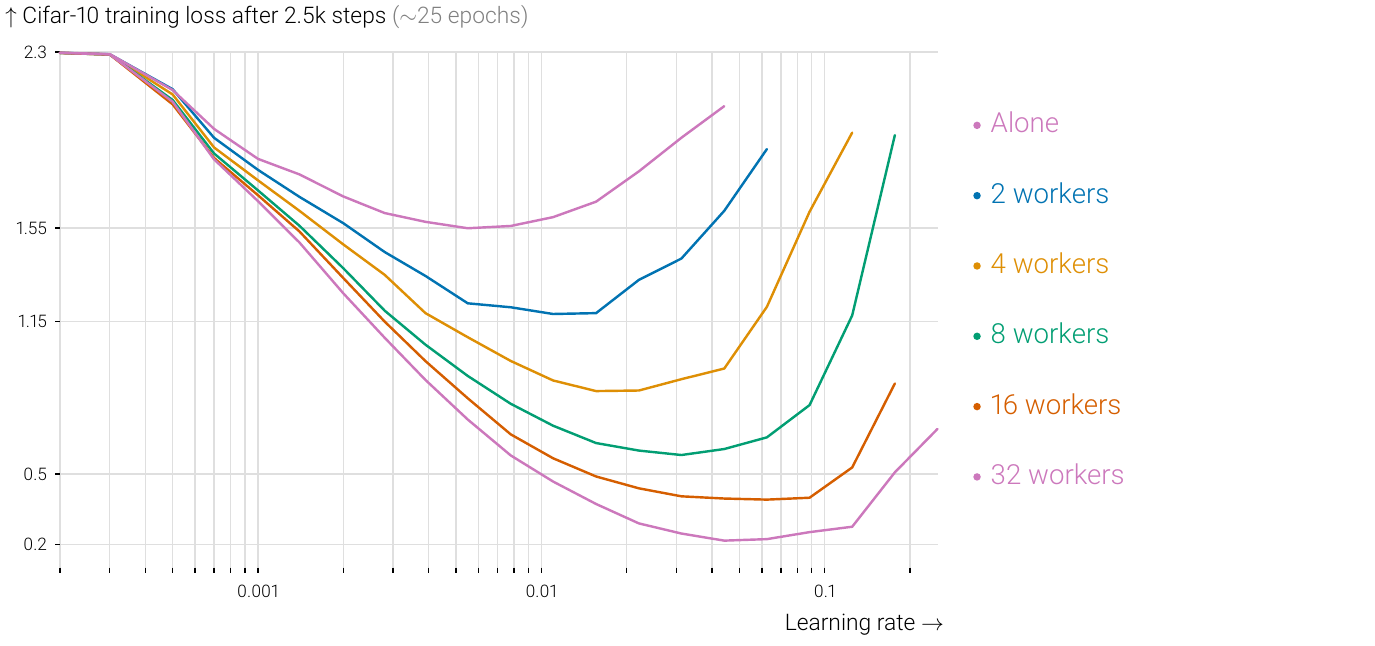}
    \caption{
        \label{apx:fig:cifar-reference-curves-fully-connected}
        Training loss reached after 2.5\kilo SGD steps with fully-connected topologies of varying size.
        Averaging with more workers speeds up convergence for fixed learning rates, but also allows larger learning rates to be used.
        This plot serves as a reference for \autoref{fig:results-cifar}, which shows similar plots for a variety of graph topologies.
    }
\end{figure}

\pagebreak
\subsection{Results on Fashion MNIST}
\label{apx:fashion}

We replicated our main experiments (Cifar-10/VGG) on another dataset and another network architecture.
We chose for the Fashion MNIST dataset~\cite{xiao2017fashion} and a simple multi-layer perceptron architecture with one hidden layer of 5000 neurons and ReLU activations.
We list the details of our experimental setup in \autoref{tab:fashion-experimental-settings}.
We varied two key parameters compared to our Cifar-10 results: we used 64 workers instead of 32, and used SGD \emph{without} momentum and \emph{without} weight decay.
Because this task is easier than Cifar-10, the initial phase where both training and test loss converge at similar rates is shorter.
We therefore consider the first 500 steps as the `initial phase', as opposed to 2500.

\begin{figure}[h]
    \includegraphics{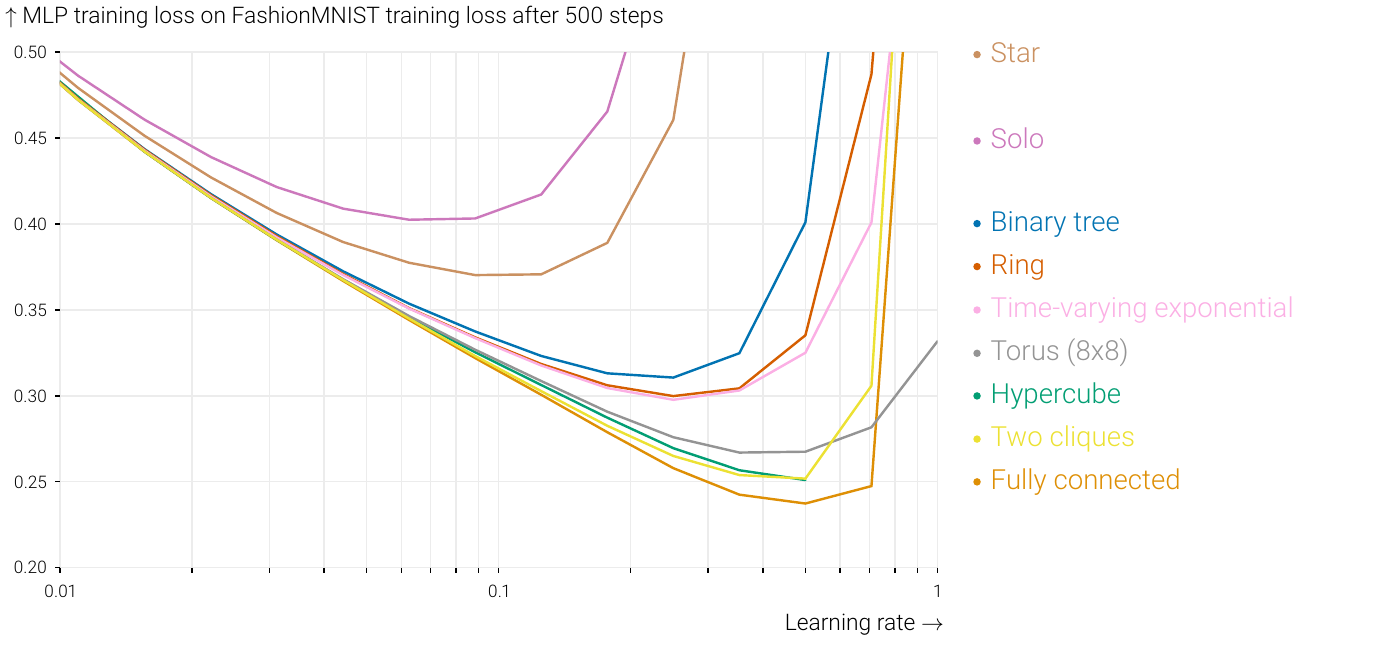}
    \caption{
        \label{apx:fig:fashion-results}
        Training loss reached after 500 SGD steps with a variety of 64-worker graph topologies.
        In all cases, averaging yields a small increase in speed for small learning rates, but a large gain over training alone comes from being able to increase the learning rate.
        While the star has a better spectral gap (0.0156) than the ring (0.0032), it performs worse, and does not allow large learning rates.
    }
\end{figure}

\begin{figure}
    \includegraphics{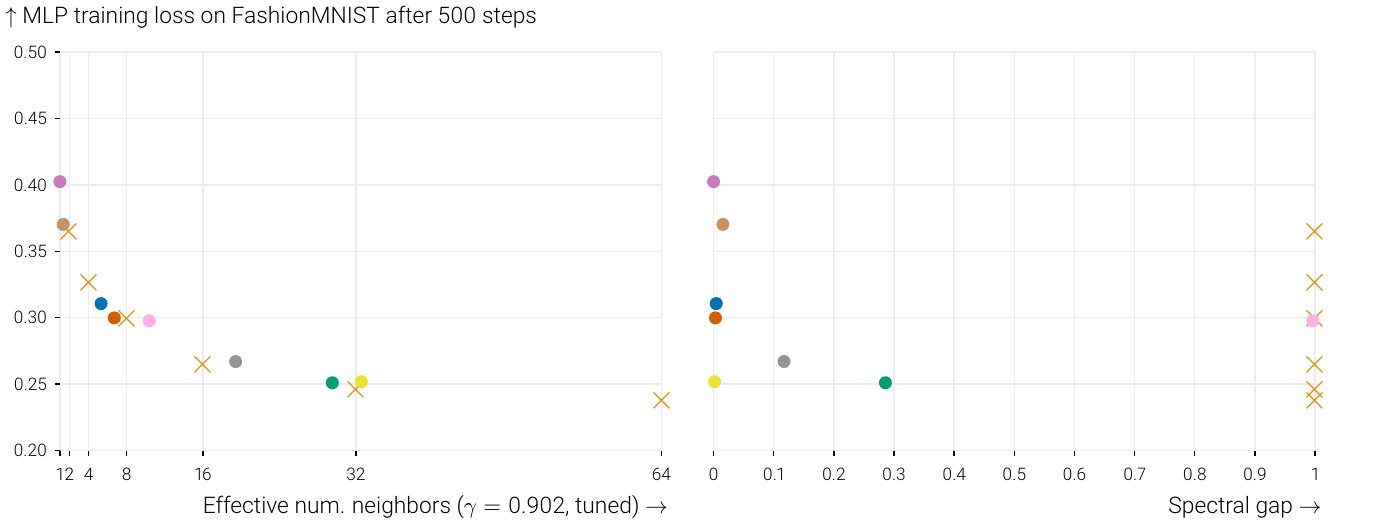}
    \centering
    \caption{
        Fashion MNIST training loss after 500 steps
        for all studied topologies with their optimal learning rates.
        Colors match \autoref{apx:fig:fashion-results}, and {\color{tab10_orange}$\times$} indicates fully-connected graphs with varying number of workers.
        After fitting a decay parameter $\decay=0.902$ that captures problem specifics, the effective number of neighbors (left) as measured by variance reduction in a random walk (like in \autoref{sec:toy-model}) explains the relative performance of these graphs much better than the spectral gap of these topologies (right).
        \label{apx:fashion_correlation_plots}
    }
\end{figure}

Figures \ref{apx:fig:fashion-results} and \ref{apx:fashion_correlation_plots} correspond to figures \ref{fig:results-cifar} and \ref{fig:correlation_plots} from the main paper.
We find that the conclusions from the paper also hold in this different experimental setting.

\begin{table}[h]
	\caption{
		Experimental settings for Fashion MNIST. Differences with Cifar-10 in \textbf{\color{tab10_red}red}.
		\label{tab:fashion-experimental-settings}%
	}
	\scriptsize%
	\begin{tabularx}{\linewidth}{lX}
		\toprule
		Dataset              & \textbf{\color{tab10_red}Fashion MNIST}~\citep{xiao2017fashion}                                                                                           \\
		Data augmentation    & \textbf{\color{tab10_red}None}                                         \\
        Data normalization   & Subtract mean 0.2860 and divide standard deviation 0.3530                          \\
		Architecture         & \textbf{\color{tab10_red}MLP} ($28\times28 \to \text{ReLU} \to 5000 \to \text{ReLU} \to 10$)                                                                              \\
		Training objective   & Cross entropy                                                                                                      \\
		Evaluation objective & Top-1 accuracy                                                                                                     \\
		\midrule
		Number of workers    & \textbf{\color{tab10_red}64} (unless otherwise specified)                                                                                                                 \\
		Topology             & Ring (unless otherwise specified) \\
		Gossip weights       & Metropolis-Hastings (1/3 for ring, $\gossipweight{ij} = 1 / (\max(n_i, n_j) + 1)$, worker $i$ has $n_i$ direct neighbors)                                                                                 \\
		Data distribution    & Identical: workers can sample from the whole dataset              \\
        Sampling    & With replacement (\iid), \emph{no} shuffled passes              \\
		\midrule
		Batch size           & 16 patches per worker                                                                                              \\
		Momentum             & \textbf{\color{tab10_red}0.0}                                                                                                     \\
		Learning rate        & Exponential grid or tuned for lowest training loss after \textbf{\color{tab10_red}500 steps}                                                                               \\
		LR decay             & None, in the initial phase of training                                                                                         \\
		LR warmup            & None                                                                \\
		\# Epochs            & 500 steps \\
		Weight decay         & \textbf{\color{tab10_red}0}                                                                                                          \\
		Normalization scheme & no normalization layers                                                                                             \\
        Exponential moving average  & $\param\atidx{t}{\text{ema}} = 0.95 \mspace{1mu} \param\atidx{t-1}{\text{ema}} + 0.05 \mspace{1mu} \param\at{t}$. This influences evaluation, not training \\
		\midrule
		Repetitions per training & Just 1 per learning rate, but experiments are very consistent across similar learning rates                                                                                                \\
		Reported metrics      & \emph{Loss after \textbf{\color{tab10_red}500} steps}:
            to reduce noise, we take two measures: (\textsc{i}) we use exponential moving average of the model parameters, and (\textsc{ii}) we fit a parametric model $\log(l)= a t + b$ to the 25 loss evaluations $(t, l)$ closest to $t=500$.
            We then evaluate this function at $t=500$.
            \\
		\bottomrule
	\end{tabularx}
\end{table}

\subsection{Heterogeneous data}
\label{apx:heterogeneous-data}

\begin{figure}
    \includegraphics{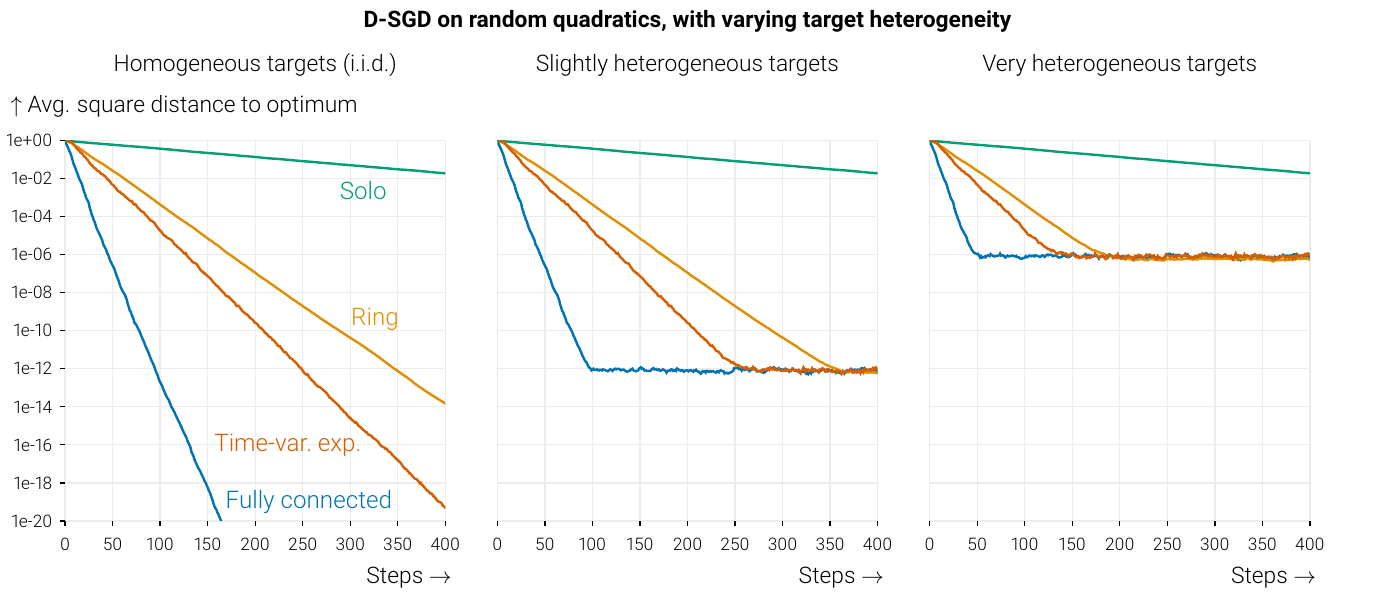}
    \centering
    \caption{
        Convergence curves on our isotropic random quadratics problem (\autoref{sec:toy-model}, with $d=100$, $n=32$), but where the optima are removed from zero as a zero-mean normal distribution with standard deviations 0, $10^{-7}$, and $10^{-3}$ respectively.
        Constant learning rates are tuned independently for each topology in the homogeneous setting.
        Heterogeneity does not affect the initial phase of training, and our insights about maximum learning rates and the quality of communication topologies hold in this regime.
        \label{apx:fig:heterogeneous-data}
    }
\end{figure}

While our experimental and theoretical data only describe the setting in which workers optimize objectives with a shared optimum,
we believe that our insights are meaningful for heterogeneous settings as well.
With heterogeneous data, we observe two regimes: in the beginning of training, when the worker's distant optima are in a similar direction, everything behaves identical to the homogeneous setting.
In this regime, our insights seem to apply directly.
Heterogeneity only plays a role later during the training, when it leads to conflicting gradient directions.
This behavior is illustrated on a toy problem in \autoref{apx:fig:heterogeneous-data}.
We run D-SGD on our isotropic quadratic toy problem ($d=100$, $n=32$), but where the optima are removed from zero as a normal distribution with standard deviations 0, $10^{-7}$, and $10^{-3}$ respectively. The (constant) learning rates are tuned for each topology in the homogeneous setting.

\subsection{The role of $\gamma$ in the experiments}

In \autoref{fig:cifar-covariance}, we optimize $\gamma$ independently for each topology, minimizing the Mean Squared Error between the normalized covariance matrix measured from checkpoints of Cifar-10 training and the covariance in a random walk with the decay parameter $\gamma$.
The bottom two rows of \autoref{apx:fig:cifar-covariance} below show how \autoref{fig:cifar-covariance} would change, if you used a $\gamma$ that is either much too low, or too high.

\begin{figure}
    \includegraphics{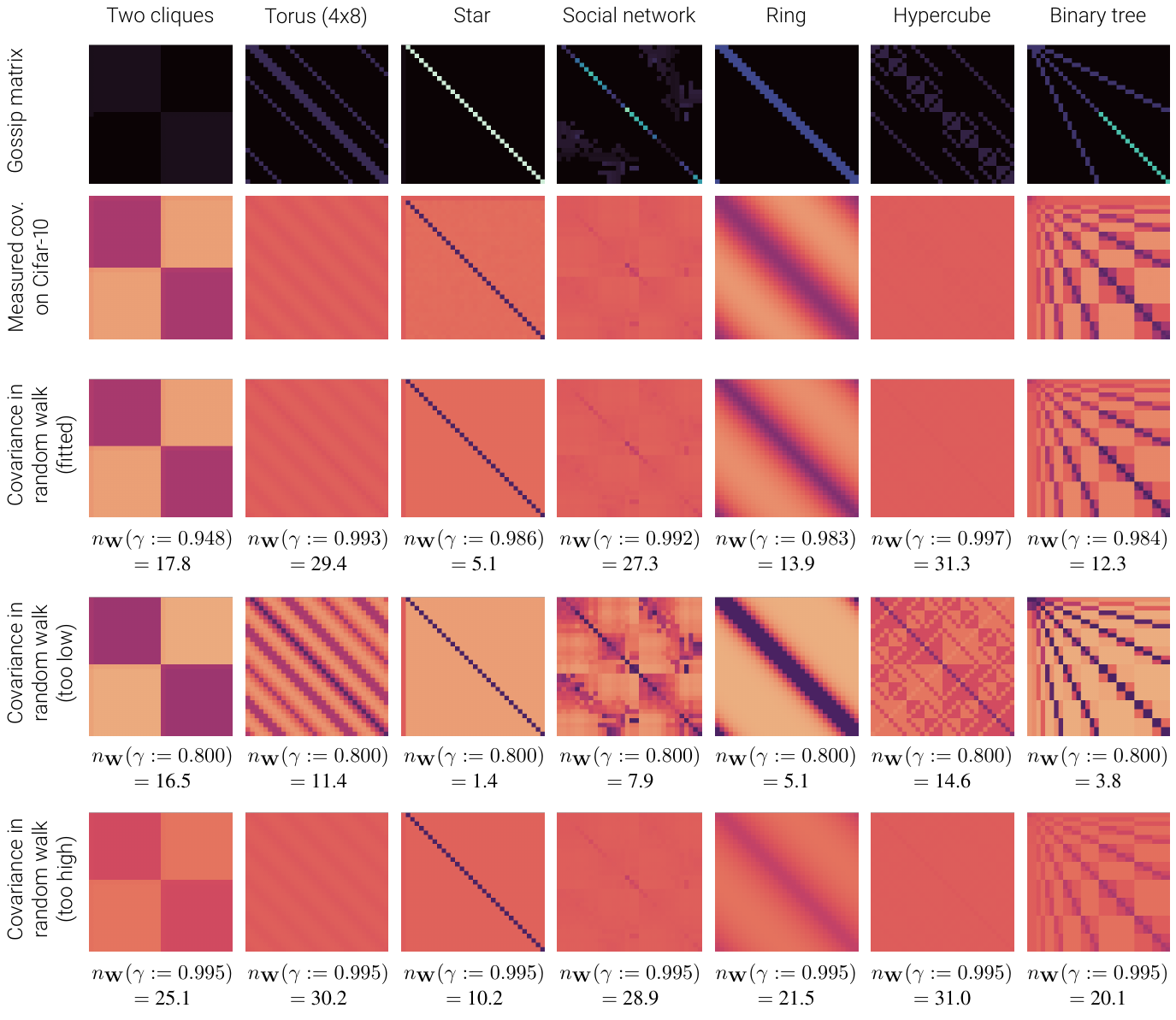}
    \centering
    \caption{
        Extension of \autoref{fig:cifar-covariance}.
        Measured covariance in Cifar-10 (second row) between workers using various graphs (top row).
        After 10 epochs, we store a checkpoint of the model and train repeatedly for 100 SGD steps, yielding 100 models for 32 workers.
        We show normalized covariance matrices between the workers.
        These are very well approximated by the covariance in the random walk process of \autoref{sec:toy-model} (third row).
        We print the fitted decay parameters and corresponding `effective number of neighbors'.
        The bottom two rows show how \autoref{fig:cifar-covariance} would change, if you used a $\gamma$ that is either much too low, or too high.
    \label{apx:fig:cifar-covariance}
    }
\end{figure}

In \autoref{fig:correlation_plots}, we choose a value of $\gamma$ (shared between all topologies) that yields a good correspondence between the performance of fully connected topologies (with 2, 4, 8, 16 and 32 workers) and the other topologies.
We opt for sharing a single $\gamma$ here, to test whether this metric could have predictive power for the quality of graphs.
\autoref{apx:fig:cifar-prediction} below shows how the figure changes if you use a value of $\gamma$ that is either much too low, or much too high.

\begin{figure}
    \includegraphics{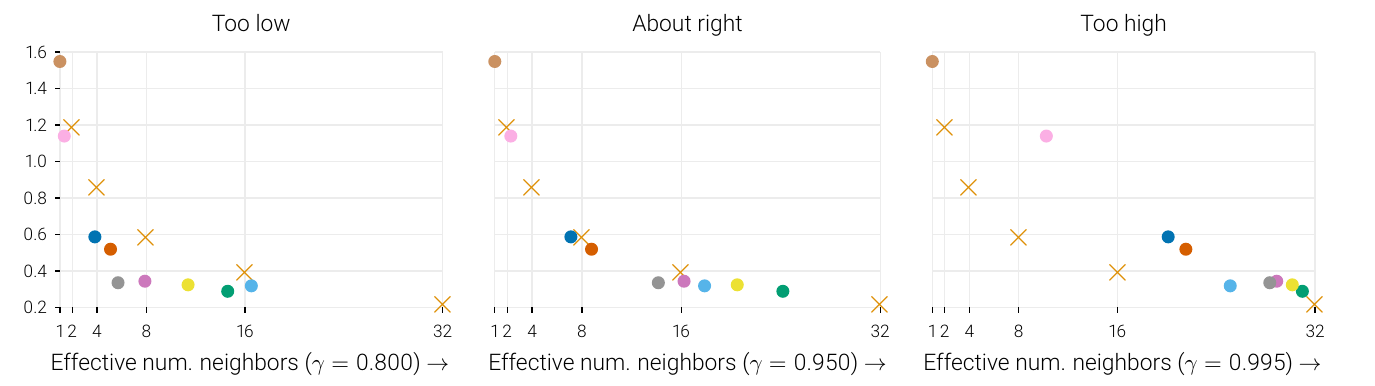}
    \centering
    \caption{
        Extension of \autoref{fig:correlation_plots}, demonstrating how the fit changes if you use a value of $\gamma$ that is either too low (left) or too high (right).
        \label{apx:fig:cifar-prediction}
    }
\end{figure}